%% file: emnlp2021.tex
\pdfoutput=1

\documentclass[11pt]{article}

\usepackage{emnlp2021}
\usepackage{subcaption}
\usepackage{times}
\usepackage{latexsym}
\usepackage{dl}
\usepackage{multirow}
\usepackage{booktabs} 
\usepackage[T1]{fontenc}

\usepackage[utf8]{inputenc}
\usepackage{algorithmic,algorithm}
\usepackage{microtype}
\newcommand{\ptrain}{p^\mathsf{train}}
\newcommand{\LL}{\mathcal{L}}
\newcommand{\ibr}{$\chi$-IBR }

\title{Distributionally Robust Multilingual Machine Translation}

\author{Chunting Zhou\thanks{\;\;Equal contribution.}$^{\;\;,1}$, Daniel Levy$^{*,2}$, Xian Li$^3$, Marjan Ghazvininejad$^3$, Graham Neubig$^1$ \\
  $^1$Language Technologies Institute, Carnegie Mellon University $\,$\\ $^2$Stanford University \\
  $^3$Facebook AI\\
     {\tt \{chuntinz, gneubig\}@cs.cmu.edu ~danilevy@stanford.edu} \\ \texttt{\{ghazvini,xianl\}@fb.com} \\
}

\begin{document}
\maketitle
\begin{abstract}
Multilingual neural machine translation (MNMT) learns to translate multiple language pairs with a single model, potentially improving both the accuracy and the memory-efficiency of deployed models.
However, the heavy data imbalance between languages hinders the model from performing uniformly across language pairs. 
In this paper, we propose a new learning objective for MNMT based on \emph{distributionally robust optimization}, which 
minimizes the worst-case expected loss over the set of language pairs. We further show how to practically optimize this objective for large translation corpora using an iterated best response scheme, which is both effective and incurs negligible additional computational cost compared to standard empirical risk minimization.
We perform extensive experiments on three sets of languages from two datasets and show that our method consistently outperforms strong baseline methods in terms of average and per-language performance under both many-to-one and one-to-many translation settings.\footnote{Our code is available at \url{https://github.com/violet-zct/fairseq-dro-mnmt}}
\end{abstract}

\input{intro}

\input{preliminaries}
\input{method}
\input{experiments}

\section{Conclusion}
We showed how to successfully apply DRO to the MNMT setting and automatically adjust the sampling distribution over language pairs resulting in sizeable improvements in performance. We posit that this approach would also be successful in other multilingual scenarios. Our work raises a few questions: (i) what are the right baseline losses? (ii) surprisingly, \ibr also improves performance on HRLs; under what circumstances does cross-lingual transfer happen and which languages does it benefit most?
We hope our work could inspire better distributionally robust learning methods for multilingual training in the future.

\section*{Acknowledgements}

This work was supported in part by a Facebook SRA Award and the NSF/Amazon Fairness in AI program under grant number 2040926.

\bibliography{anthology}
\bibliographystyle{acl_natbib}

\clearpage
\newpage
\appendix

\label{sec:appendix}

\input{primal_dual}
\input{appendix}
\end{document}

%% file: intro.tex
\section{Introduction}
Multilingual methods that process multiple languages with one single model have gained favor across a variety of NLP tasks~\citep{firat2016multi,ha2016toward,johnson2017google,kenton2019bert,aharoni2019massively,conneau2020unsupervised} because (1) training and deployment of one multilingual model is more computationally efficient compared to maintaining one model for each language~\citep{arivazhagan2019massively}, (2) training multilingually can improve accuracy, particularly for low-resource languages (LRLs)~\citep{zoph2016transfer,neubig2018rapid,pires2019multilingual}.

\begin{figure}[t]
    \centering
      \includegraphics[width=0.35\textwidth]{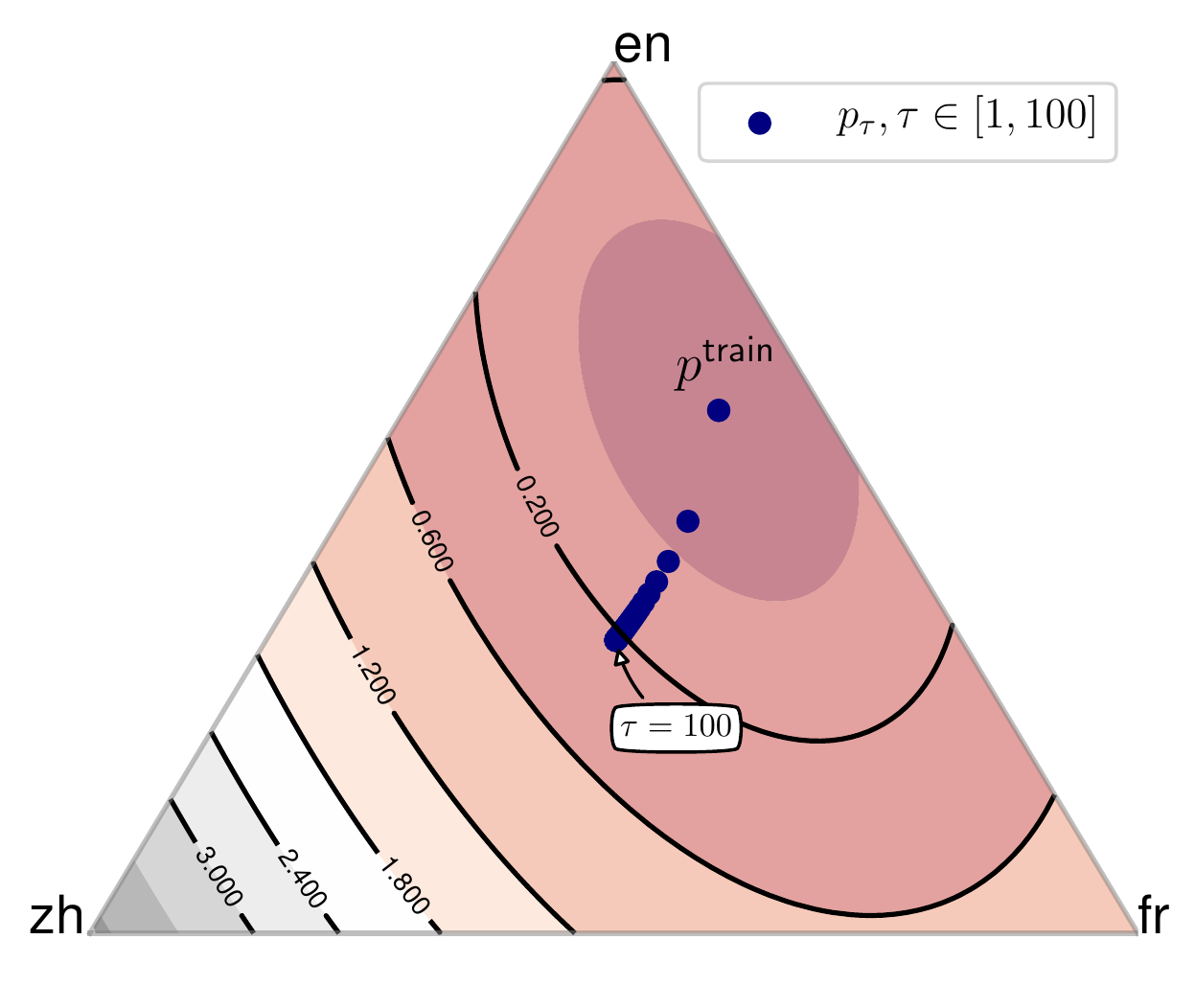}
      \vspace{-2mm}
    \caption{Illustration of different training distributions where the training distribution of the three languages \texttt{fr}, \texttt{zh} and \texttt{en} is $(0.1, 0.3, 0.6)$. Contours represent different radii of the $\chi^2$-ball around $\ptrain$.
    The blue points are the tempered distributions described in \S\ref{sec:mmt}.}
    \label{fig:intro}
    \vspace{-4mm}
\end{figure}
However, in multilingual training, the amount and type 
of training data available varies greatly across languages.
Because most models are trained using empirical risk minimization (ERM), which minimizes the average training loss on the training set, 
high-resource languages (HRLs) with large amounts of data contribute to the majority of the training objective. When model capacity is limited, this results in trade-offs or decreased performance on some languages, particularly LRLs~\citep{arivazhagan2019massively,wang2020negative,wang2021gradient}. 
To better control this trade-off, a common practice is to balance the training distribution by heuristic oversampling of LRLs~\citep{johnson2017google,neubig2018rapid,arivazhagan2019massively}. 

Although simple data balancing can improve the performance on LRLs significantly, it is far from optimal. First, the sampling hyperparameters need to be adjusted for different datasets. Second, the use of simple heuristics does not consider the inherent level of difficulty in learning each language, the similarity between languages in the multilingual dataset,  and other factors that affect cross-lingual transfer.
Because of this, previous work has indicated the importance of learning strategies that are explicitly tailored for each multilingual learning scenario~\citep{wang2020balancing}. 

In this paper, we propose a new learning procedure for multilingual translation that automatically adjusts the training distribution of different languages using distributionally robust optimization (DRO)~\citep{ben2013robust,duchi2016statistics}.
In constrast to ERM, DRO casts learning as a game between the learner and an adversary, where the learner picks a model while the adversary picks the hardest data distribution for that model within an \emph{uncertainty set} $\mc{Q}$ of potential distributions we wish to perform well on (which typically contains the training distribution $P_0$).

We first demonstrate how to apply DRO to multilingual training by letting the adversary choose the relative weights of individual language pairs in the training objective. 
However, we empirically find that naively applying existing methods to multilingual learning yields inferior results to ERM, mostly because (1) standard DRO objectives tend to be overly conservative and only take into account language pairs with very large losses and (2) existing optimization algorithms for DRO essentially reweigh the gradients of examples in a mini-batch, which implicitly changes the scale of the learning rates. This hurts modern NLP models 
like Transformers~\citep{vaswani2017attention} that are highly sensitive to learning rate schedules. 

To remedy this, we propose both a novel training objective and a corresponding optimization algorithm amenable to the multilingual setting. Our objective is a variation on Group DRO of~\citeauthor{sagawa2019distributionally} with a more flexible uncertainty set, parameterized by the $\cs$-divergence. To efficiently solve the min-max game, we propose an \emph{iterated best response} scheme that, at each epoch, re-samples the training data according to the worst weighting for the current model parameters, and then runs ERM training on the re-sampled dataset. Our method---which we refer to as \ibr---incurs negligible additional computational cost compared to ERM.

While this method applies to essentially any multilingual task, we specifically demonstrate its benefit on three sets of language pairs from two multilingual machine translation datasets. 
We experimentally test these choices by comparing several objectives and optimization algorithms, and results show that our method consistently outperforms existing DRO 
procedures and various strong baselines.


%% file: preliminaries.tex
\section{Preliminaries}\label{sec:prelim}

\paragraph{Notation.}

Throughout this paper, $n$ denotes the training set size and $d$ the number of parameters of the model. For $m \in \N$, $\Delta^m$ denotes the $m$-dimensional simplex, i.e.~$\Delta^m \defeq \crl{q \in \R^m, q_i \ge 0\mbox{~and~} \sum_i q_i = 1}$. The data lies in $\mc{X}\times\mc{Y}$ where $(\vx, \vy) \in \mc{X}\times\mc{Y}$ consists of a source and target sentence pair with $\vx = (\vx_1, \ldots, \vx_{L_{\vx}})$ and $\vy = (\vy_1, \ldots, \vy_{L_{\vy}})$. The function $\ell:(\mc{X}\times\mc{Y})\times\R^d \to \R$ refers to the loss. 
We consider maximum-likelihood estimation, i.e.~for a target sentence $\vy$ with $L_\vy$ tokens, we define
\vspace{-3mm}
\begin{equation}\label{eq:loss-function}
\small
    \ell(\vx, \vy; \theta) = -\frac{1}{L_\vy}\sum_{i=1}^{L_\vy} \log p(\vy_i|\vx, \vy_{<i}; \theta)\nonumber
\vspace{-1mm}
\end{equation}

\subsection{Multilingual Machine Translation}
\label{sec:mmt}


In contrast to bilingual machine translation, which translates from a single source language $S$ to a target language $T$, multilingual neural MT (MNMT) learns a single model to translate between $N$ language pairs $\crl{\prn{S_1,T_1}, \ldots \prn{S_N, T_N}}$
. 
The training data $D_{\mathrm{train}}$ is the concatenation of the $N$ parallel datasets, i.e.~$D_{\mathrm{train}} = [D_1; D_2, \cdots; D_N]$. We can then define the probability over each language pair $\ptrain \in \Delta^N$ as $\ptrain_i = \tfrac{\abs{D_i}}{\sum_j \abs{D_j}}$.
We now describe two common training objectives for MNMT.
\paragraph{Empirical Risk Minimization (ERM).} The simplest and most common approach for MNMT is to minimize the empirical loss over data points, which we will refer to as ERM. More precisely, we define the average loss on a parallel dataset $D_i$ as
\begin{equation}
\nonumber
\small
\vspace{-1mm}
    \LL(\theta;D_i) \defeq \frac{1}{\abs{D_i}}\sum_{(\vx, \vy) \in D_i} \ell(\vx, \vy;\theta).
    \vspace{-1mm}
\end{equation}
ERM for multilingual models then corresponds to simply minimizing the loss over $D$, i.e.~over all the aggregated parallel sentence pairs. This yields
\begin{equation*}
\vspace{-1mm}
    \what{\theta}_n^\mathsf{ERM} \in \argmin_{\theta} \LL(\theta;D) = \sum_{i\le N} \ptrain_i \LL(\theta;D_i).
    \vspace{-1mm}
\end{equation*}
Classical results in learning theory guarantee that, under mild assumptions, 
as $n$ goes to infinity, 
$\what{\theta}_n^\mathsf{ERM}$ will show good performance on test sets with the same distribution as $D$. However, this does not guarantee that our model will perform adequately on individual parallel datasets. To remedy this issue, several works propose varying the sampling distribution---or equivalently the weighting of the parallel datasets in ERM---in order to encourage more uniform performance across language pairs.


\paragraph{Weighted Risk Minimization and Sampling Strategies.}
The amount of training data can vary significantly across language pairs. As a result, in ERM training---i.e.~optimizing for the average loss across sentence pairs---HRLs contribute most of the objective, resulting in poor performance on LRLs. 
Balancing the objective---or equivalently, the usage of training data---between HRLs and LRLs is important to maintain good performance across all languages~\citep{kenton2019bert,arivazhagan2019massively}. 
A commonly adopted approach in multilingual training is \emph{temperature}-based sampling~\citep{arivazhagan2019massively,conneau2020unsupervised} where the probability of sampling data from $D_i$ is proportional to its data size exponentiated by a temperature term $\tau$, i.e.~$p_{\tau,i} = \tfrac{\abs{D_i}^{1/\tau}}{\sum_j \abs{D_j}^{1/\tau}}$ (referred to as ERM with $\tau$ in \S\ref{sec:experiments}). This is equivalent to optimizing the re-weighted objective
\begin{equation}
\nonumber
\small
\LL_\tau(\theta;D) = \sum_{i\le N}p_{\tau,i}\LL(\theta;D_i).
    \vspace{-1mm}
\end{equation}
As a result, $\tau=1$ corresponds to ERM where most of the contribution comes from the HRLs and $\tau=\infty$ corresponds to sampling language pairs uniformly at random
, i.e.~
with data from LRLs being over-sampled. This approach comes with three major drawbacks 
(1) $\tau$ is an extra hyper-parameter that requires tuning for each MNMT instance to balance the performance across both HRLs and LRLs, (2) this heuristic sampling method does not consider the training dynamics of each language and how the optimal sampling distribution might evolve during the training process and (3) the parameterization of $p_\tau$ is not only very constrained (essentially one degree of freedom), it is also only a function of the quantity of training data, which is too rigid to achieve the desired performance.

To resolve some of the above issues, the recently proposed MultiDDS \citep{wang2020balancing} uses a gradient-based meta-learning approach to learn the sampling distribution over language pairs to maximize gradient similarity with a multilingual development set. However, due to the necessity to calculate and store extra gradients, their approach comes at an increased computational and memory cost. In contrast, \ibr enjoys the same computational complexity as ERM, and as we show in experiments it also largely outperforms MultiDDS.





\subsection{Distributionally Robust Optimization}



In contrast to ERM and related sampling strategies which optimize for a fixed training distribution, DRO aims to find a model $\theta$ that performs well on an entire collection of potential test distributions $\mc{Q}$ (the \emph{uncertainty set}). Formally, we wish to
\begin{equation}\label{eq:dro}
    \minimize_{\theta}\sup_{Q \in \mc{Q}}
    \E_{(\vx, \vy)\sim Q}\brk{\ell(\vx, \vy; \theta)}.
\end{equation}

Originating from operations research~\cite{DelageYe10, ben2013robust, Ben-TalHeWaMeRe13, BertsimasGuKa18}, DRO is a promising way to tackle robustness in a variety of machine learning and NLP problems~\cite{HashimotoSrNaLi18,oren2019distributionally, LevyCaDuSi20}
.

We present here a recent variant, Group DRO, developed by \citet{sagawa2019distributionally} which incorporates additional information about the data distribution to define more meaningful uncertainty sets. Abstractly, this method assumes a collection of distributions over subpopulations $\crl{P_g}_{g\in\mc{G}}$ such that the training distribution is a mixture of these subpopulations.
Importantly, it assumes that this \emph{group structure} is observed. The Group DRO 
objective then minimizes the worst-case loss over these groups, which is equivalent to~\eqref{eq:dro} with $\mc{Q} = \crl{\sum_{g\in\mc{G}} q_gP_g: q\in\Delta^{\abs{\mc{G}}}}$, or equivalently, all possible mixtures of the distributions over subpopulations. 
In MNMT, the $N$ language pairs at our disposal naturally correspond to groups; thus the Group DRO objective can be defined as
\begin{equation}\label{eq:gdro}
\vspace{-2mm}
    \LL^{\mathsf{GDRO}}(\theta;D) = \max_{i\in \brk{N}} \LL(\theta;D_i).
\vspace{-1mm}
\end{equation}
In other words, Group DRO wishes to find a model $\theta$ that performs well for the \emph{worst} language pair. \citet{oren2019distributionally} propose a related but less conservative objective, CVaR-Group DRO at level $\alpha\in\brk{0, 1}$ which, considers instead the average of the $\ceil{\alpha N}$ largest group losses.

%% file: method.tex
\section{Methods for Distributionally Robust Multilingual Learning}


As we previewed in \S\ref{sec:prelim}, Group DRO is a natural objective for the multilingual setting. However, in experiments we found that naively applying existing DRO objectives fails to achieve performance on par with strong baselines, often improving results on 
language pairs with high losses
but sacrificing too much performance overall. Our main contribution is showing how to successfully apply DRO to the MNMT setting, and to the best of our knowledge, our work is the first to do so. To that end, our methodological contributions are two-fold: (i) we first describe the shortcomings of the Group DRO objective~\eqref{eq:gdro}, then propose a related training criterion that addresses these issues, (ii) we describe an optimization algorithm to solve the min-max optimization problem that is amenable to the MNMT setting.


\subsection{$\cs$-Group DRO} 

\paragraph{Shortcomings of Group DRO.} A weakness of the objective~\eqref{eq:gdro} is that apart from the language pair with largest loss, the objective does not take into account the value of the loss on the other language pairs. To illustrate this, consider this example with $N=3$ language pairs and suppose that there exists two parameters $\theta_1$ and $\theta_2$ with the following loss:
\vspace{-2mm}

{\small 
\begin{align*}
    \LL(\theta_1;D_1) = 0.1, \, \LL(\theta_1;D_2) = 0.1, \, \LL(\theta_1; D_3) = 1.1 \\
    \LL(\theta_2;D_1) = 1.0, \, \LL(\theta_2; D_2) = 1.0, \, \LL(\theta_2; D_3) = 1.0
\end{align*}}
We have that $\LL^{\mathsf{GDRO}}(\theta_1;D) = 1.1$ but $\LL^{\mathsf{GDRO}}(\theta_2;D) = 1.0$. Consequently, the Group DRO objective will prefer $\theta_2$ to $\theta_1$ while clearly one would pick $\theta_1$ over $\theta_2$ in most practical cases. The aforementioned CVaR-Group DRO also exhibits this behavior and ignores the values of the language pairs not in the largest $\alpha$-fraction.



To address this issue, let us rewrite the objective
\begin{equation*}
    \LL^{\mathsf{GDRO}}(\theta;D) = \max_{q\in\Delta^N}\sum_{i\le N}q_i\LL(\theta;D_i),
\end{equation*}
where the equality holds because the 
optimal weighting just puts all the mass on the language with largest loss. A natural way to make the objective less conservative is to instead take the maximum over a \emph{subset} of the simplex $\mc{U} \subset \Delta^N$. This leads to the following objective
\begin{equation}\label{eq:gen-gdro}
    \LL^{\mc{U}}(\theta;D) = \sup_{q\in\mc{U}}\sum_{i\le N}q_i\LL(\theta;D_i).
\end{equation}

Different choices of $\mc{U}$ will yield different objectives with different robustness properties. Note that this is a general formulation as $\mc{U} = \crl{\ptrain}$ reduces to the ERM objective
, while $\mc{U}_\alpha = \crl{q: \norms{q / \ptrain}_\infty \le 1/\alpha}$ corresponds to the CVaR-Group DRO of~\citet{oren2019distributionally}. 
We would like to choose $\mc{U}$ such that optimizing this objective results in models with better performance on language pairs with large losses (typically LRLs) without significant degradation of average performance.


To this end, we turn to a common and flexible choice for $\mc{U}$: $f$-divergence balls~\cite{Csiszar67} of radius $\rho > 0$ around $\ptrain$, namely
\begin{equation}
    \mc{U}^f_{\rho} \defeq \crl{q: D_f(q, \ptrain) \le \rho},
\end{equation}
where $D_f(q, p) \defeq \sum_{i\le N}p_i f(q_i/p_i)$. In particular, we propose using the $\cs$-divergence which corresponds to $f(t) = \tfrac{1}{2}(t-1)^2$ and define $\cs(q, p) = \tfrac{1}{2}\sum_i p_i(q_i/p_i - 1)^2$ with its corresponding uncertainty set $\mc{U}^{\cs}_{\rho}$. The $\cs$-divergence has a long history~\cite{Ben-TalHeWaMeRe13} and previous work shows that minimizing the robust loss with the $\cs$-uncertainty set enjoys favorable statistical properties such as optimally trading-off bias and variance~\cite{DuchiNa19} or guaranteeing robustness and fairness~\cite{HashimotoSrNaLi18, DuchiNa20}. We refer to the objective $\LL^{\mc{U}^{\cs}_{\rho}}$ as the $\cs$-Group DRO. Going back to the toy example, setting $\rho = 0.1$, yields that $\LL^{\mc{U}^{\cs}_{\rho}}(\theta_1;D) = 0.64$ while $\LL^{\mc{U}^{\cs}_{\rho}}(\theta_2;D) = 1.0$. With the $\cs$-uncertainty set, the objective rightly prefers $\theta_1$ to $\theta_2$ and takes into account all the losses and not only the largest. We further confirm these intuitions and show in \S\ref{sec:experiments} and \S\ref{sec:analysis} that this is a suitable choice of uncertainty set for MNMT. 



\subsection{Optimization algorithm}
\label{sec:opt:alg}
\paragraph{Desiderata of the optimization algorithm.} Minimizing the objective~\eqref{eq:gen-gdro} effectively corresponds to a min-max optimization problem. Even in the relatively simple convex-concave setting, these are generally harder to solve than convex minimization problems. Recall that we want to
\begin{equation}
    \minimize_\theta \sup_{q:\cs(q, \ptrain) \le \rho} \sum_{i\le N}q_i \LL(\theta;D_i).
\end{equation}

Due to the architectures we consider in MNMT, we wish for an algorithm that effectively changes the sampling distribution over mini-batches of data instead of importance-weighting the gradients. Indeed, standard MT architectures such as the Transformer \citep{vaswani2017attention} are extremely sensitive to learning rate schedules and we empirically observe that importance-weighting the gradients result in poor performance. 

The canonical way to solve min-max problem is via primal-dual methods (PD)~\cite{NemirovskiJuLaSh09} (see background in Appendix~\ref{app:primal-dual}), where at each step $t$, one keeps two vectors $(\theta_t, q_t)$ and alternates between a gradient descent step on $\theta_t$ and a gradient ascent step on $q_t$. One can perform these updates efficiently as they only require \emph{unbiased stochastic gradient estimate} of the loss w.r.t.~$\theta_t$ and $q_t$. To obtain unbiased gradient estimate of the loss w.r.t.~$\theta_t$, one either has to, at each step, sample a mini-batch of examples from $\mathsf{Multinomial}(q_t)$ and return the gradient of the loss or sample a mini-batch from $\mathsf{Multinomial}(\ptrain)$ and importance-weight the gradient. 

As previously mentioned, the latter is not suitable for Transformer-type architectures. The former option is not ideal as it is more convenient for an algorithm to decide the sequence of mini-batches every epoch rather than every optimizer step as this integrates much more smoothly with data loaders in deep learning frameworks, especially when doing distributed training. As a result, we posit that primal-dual algorithms are not an adequate choice for optimizing DRO-type objectives in our setting. We further discuss this point in \S\ref{sec:analysis}.



To circumvent this issue, we consider a different optimization algorithm which we refer to as \emph{iterated best response} (IBR), where, instead of doing a single gradient descent and ascent step, we iterate between (approximately) solving the maximization (resp. minimization) on $q$ (resp. $\theta$), while keeping $\theta_t$ (resp. $q_t$) fixed. This is similar in spirit to algorithms in the game theory literature where individuals play the optimal strategy (best response) assuming everyone else's strategies remain constant. Under some mild assumptions, this procedure converges to the equilibrium of the game~\cite{Roughgarden16}. 
Formally, we alternate between
\begin{align}
\small
\vspace{-2mm}
        \theta^{t+1} & \leftarrow \argmin_{\theta} \sum_i q^{t}_i \LL(\theta;D_i)\label{eq:update-theta}\\
        q^{t+1} & \leftarrow \argmax_{q:\cs(q, p^\msf{train})\le \rho}
        \sum_i q_i \LL(\theta^{t+1};D_i)\label{eq:update-q}.
    \vspace{-2mm}
\end{align}
\input{iterated_best_response}
\vspace{-5mm}
\paragraph{Practical implementation.} As we show in Appendix~\ref{app:best-response}, given the values of the loss, the $q$-update~\eqref{eq:update-q} is computed to accuracy $\epsilon$ in $O(N\log(1/\epsilon))$ steps. Indeed, by taking the dual~\cite{BoydVa04} of~\eqref{eq:update-q}, we transform the $N$-dimensional problem into a one-dimensional root-finding procedure over the dual variable which we efficiently solve with a bisection. We provide the details in Appendix~\ref{app:best-response}. Note that computation cost is negligible compared to computing the gradient of the loss. We implement the $\theta$-update of~\eqref{eq:update-theta} by running a training epoch on a \emph{re-sampling} of the training set $D$ according to $q^{t+1}$. 

To compute the loss values $\crl{\LL(\theta_t;D_i)}_{i\le N}$, necessary to perform the $q$-update, one needs to compute the loss $\ell(\vx, \vy;\theta_t)$ for every single example $(\vx, \vy)\in D$. This is prohibitively expensive to compute at every epoch. To avoid this, we keep track of the (approximate) historical values of the token-level loss $\hat{\LL}_k$ on each language pair $k$ with an exponential moving average (EMA) (see line~14 
in Algorithm~\ref{alg:iterated-best-response}).
We precisely describe our implementation in Algorithm~\ref{alg:iterated-best-response}. We see that it respects our desiderata and comes at no computational cost. In \S\ref{sec:analysis}, we compare primal-dual and iterated best response for various uncertainty sets.

\vspace{-1mm}
\paragraph{Subtracting the baseline.}   \citeauthor{oren2019distributionally} propose subtracting a per-group scalar---which we refer to as a baseline---to each group loss before taking the maximum over $q$. They learn this baseline using a generative bi-gram model on each group. Recall that $\what{\theta}^{\mathsf{ERM}}$ is the parameter we obtain when we minimize the average loss and define $\what{\theta}^\tau$ when optimizing $\LL_\tau$. In this work, we propose using $b_i = \LL(\what{\theta}^{\mathsf{ERM}};D_i)$ or $b_i = \LL(\what{\theta}^\tau;D_i)$. Intuitively, the baseline corresponds to the minimum performance we wish for our model on the given group and as such the loss obtained with ERM and its temperature variants are natural candidates. We show in \S\ref{sec:analysis} that these yield significant improvement and conveniently make our method more robust to the choice of $\rho$. We leave different (potentially learned) choices of baseline to future work.

%% file: iterated_best_response.tex
\begin{algorithm}[t]
    \small
	\caption{Iterated Best Response}
	\begin{algorithmic}[1]
		\STATE \textbf{Input:} $N$ parallel datasets $D_1, \ldots, D_N$, radius of the uncertainty set $\rho$, number of epochs $T$, learning rate schedule $\crl{\eta_{t, j}}_{t\le T, j \le n}$, baseline loss $\crl{b_i}_{i\le N}$, EMA parameter $\lambda \in \brk{0, 1}$.
		
		\STATE Set $\ptrain_i \leftarrow \abs{D_i} / (\sum_j \abs{D_j})$.
		\STATE Set $q^0 \leftarrow \ptrain$ and $\what{\LL}_i \leftarrow 0$ for $i \in \brk{N}$.
		\STATE Initialize $\theta^0$.
		\FOR {$t=0,\ldots,T-1$}
		    \STATE\COMMENT{Construction of $D(q^t)$}
		    \STATE Set $n(q^t)_i \leftarrow \ceil{N\cdot q^t_i}$.
		    \STATE Sample $n(q^t)_i$ data points from $D_i$ and add them to $D(q^t)$ for $i \in \brk{N}$.
		    
		  \STATE $D(q^t) \leftarrow \mathsf{Shuffle}(D(q^t))$.
		  \STATE\COMMENT{$\theta$- and $\what{\LL}$-update}
		    \FOR{$(\vx_j, \vy_j) \in D(q^t)$}
		        \STATE Let $k$ be the language pair of $(\vx_j, \vy_j)$.
		        \STATE $\theta^{t, j} \leftarrow \theta^{t, j-1} - \eta_{t, j}\nabla \ell(\vx_j, \vy_j;\theta^{t, j-1})$.
		        \STATE{$\what{\LL}_{k} \leftarrow \lambda \cdot  \ell(\vx_j, \vy_j;\theta^{t, j-1}) + 
		        (1-\lambda) \cdot \what{\LL}_{k}.$} \label{line:ema}
		    \ENDFOR
		    \STATE\COMMENT{$q$-update}
		        \STATE \quad\; $q^{t+1} \leftarrow \argmax_{q:\cs(q, \ptrain)}
		        \sum_{i\le N} q_i\prn*{\what{\LL}_i - b_i}$
		    \STATE $\theta^{t+1, 0} \leftarrow \theta^{t, \abs{D(q^t)}}$.
		\ENDFOR
		\STATE Return $\theta^{T, 0}$.
	\end{algorithmic}
	\label{alg:iterated-best-response}
\end{algorithm}



%% file: experiments.tex
\begin{table*}[t]
\small
    \centering
    \begin{tabular}{l|l | cccccccc c}
    \toprule
        & \multirow{2}{*}{\textbf{Method}} & \textbf{aze} & \textbf{bel}	& \textbf{glg} & \textbf{slk} & \textbf{tur} & \textbf{rus} & \textbf{por} & \textbf{ces}	& \multirow{2}{*}{\textbf{Avg}}\\
        & &  0.004 & 0.006 & 0.013 & 0.081 & 0.240 & 0.274 & 0.243 & 0.136 & \\
    \midrule
        \multirow{3}{*}{any$\rightarrow$en} & ERM ($\tau$=1) & 14.11 & \underline{20.14} & \textbf{31.94} & 32.47 & \underline{27.18} & \underline{25.68} & \underline{45.26} & \underline{30.26} & 28.38 \\
        & MultiDDS & \textbf{14.97} & \textbf{20.60} & 31.70 & \underline{32.54} & 26.56 & 25.40 & 44.67 & 29.79 & 28.28 \\
        & \ibr & \underline{14.68} & 19.98 & \underline{31.89} & \textbf{33.16} & \textbf{27.76} & \textbf{26.08} & \textbf{45.33} & \textbf{30.76} & \textbf{28.71} \\
        \midrule
        \multirow{3}{*}{en$\rightarrow$any} & ERM ($\tau$=1) & 7.61 & 12.68 & 24.79 & \underline{25.24} & \underline{16.79} & \underline{20.80} & \underline{40.84} & \underline{22.93} & 21.46 \\
        & MultiDDS & \underline{8.04} & \textbf{15.04} & \textbf{26.60} & 25.19 & 16.32 & 20.44 & 40.47 & \underline{22.93} & 21.88 \\
         & \ibr & \textbf{8.49} & \underline{13.84} & \underline{25.77} & \textbf{26.09} & \textbf{16.78} & \textbf{21.59} & \textbf{41.38} & \textbf{23.70} & \textbf{22.21} \\
    \midrule
    & \multirow{2}{*}{\textbf{Method}} & \textbf{bos} & \textbf{mar} & \textbf{hin} & \textbf{mkd} & \textbf{ell} & \textbf{bul} & \textbf{fra} & \textbf{kor}	& \multirow{2}{*}{\textbf{Avg}} \\
    & & 0.007 & 0.017 & 0.033 & 0.045 & 0.237 & 0.308 & 0.340 & 0.363 & \\
    \midrule
    \multirow{3}{*}{any$\rightarrow$en} & ERM ($\tau$=1) & 24.79 & 12.12 & 23.76 & 34.32 & \underline{39.17} & \underline{40.17} & \textbf{41.08} & \underline{20.19} & 29.45 \\
    & MultiDDS & \textbf{26.39} & \textbf{12.62} & \textbf{24.62} & \textbf{34.65} & 38.46 & 39.71 & 40.60 & 19.46 & 29.56 \\
     & \ibr & \underline{25.12} & \underline{12.52} & \underline{24.42} & \underline{34.47} & \textbf{39.42} & \textbf{40.24} & \underline{40.98} & \textbf{20.72} & \textbf{29.74}  \\
     \midrule
     \multirow{3}{*}{en$\rightarrow$any} & ERM ($\tau$=1) & 16.29 & \underline{5.59} & 16.83 & \underline{26.42} & \underline{33.22} & \underline{35.79} & \underline{39.68} & \underline{9.09} & 22.86 \\
     & MultiDDS & \textbf{17.96} & \textbf{5.61} & \textbf{17.44} & 25.98 & 32.72 & 35.28 & 39.57 & 8.96 & 22.94 \\
     & \ibr & \underline{17.33} & \underline{5.59} & \underline{16.90} & \textbf{28.02} & \textbf{33.82} & \textbf{36.37} & \textbf{40.35} & \textbf{9.13} & \textbf{23.44} \\
    \bottomrule
    \end{tabular}
    \caption{BLEU scores of the best ERM model (among $\tau$=1/5/100, $\tau=5/100$ are significantly worse than $\tau=1$, thus we omit these results), MultiDDS~\citep{wang2020balancing} and our approach on the test sets of the TED dataset. Bold (resp. underlined) values indicate the best (resp. second best) performance for each language pair. Values under the language codes are the proportion of the language in the training data.}
    \label{tab:ted:full}
    \vspace{-3mm}
\end{table*}

\begin{table*}[t]
\small
    \centering
    \begin{tabular}{l| cccc c| cccc c}
    \toprule
        \multirow{2}{*}{\textbf{Method}} & \multicolumn{5}{c|}{any$\rightarrow$en} & \multicolumn{5}{c}{en$\rightarrow$any}  \\
       & \textbf{deu} & \textbf{fra} & \textbf{tam} & \textbf{tur} & \textbf{Avg} & \textbf{deu} & \textbf{fra} & \textbf{tam} & \textbf{tur} & \textbf{Avg} \\
    \midrule
        ERM ($\tau$=1) &  \textbf{29.98} & 30.32 & 15.81 & 19.85 & 23.99 & \textbf{23.82} & 33.09 & 9.28 & 13.29 & 19.87 \\
        ERM ($\tau$=5) & 29.25 & \underline{31.60} & \underline{16.31} & 21.89 & \underline{24.76} & 22.67 & \underline{32.36} & 10.04 & \underline{16.09} & \underline{20.29} \\
        ERM ($\tau$=100) & 28.75 & 30.71 & 15.80 & \textbf{22.44} & 24.43 & 22.02 & 31.65 & \underline{10.41} & \textbf{16.44} & 20.13 \\
        MultiDDS &  29.31 & 31.41 & 16.12 & 21.43 & 24.57 & 22.99 & 31.55 & 10.09 & 14.51 & 19.79 \\
        \ibr & \underline{29.67} & \textbf{31.75} & \textbf{16.48} & \underline{22.33} & \textbf{25.06} & \underline{23.45} & \textbf{33.16} & \textbf{10.73} & 15.53 & \textbf{20.72} \\
    \bottomrule
    
    \end{tabular}
    \caption{BLEU scores of the ERM ($\tau$=1/5/100), MultiDDS and our method on the test sets of the WMT dataset. The ratios of training data of de, fr, ta and tr are (0.499, 0.359, 0.102, 0.039). }
    \label{tab:wmt:full}
    \vspace{-3mm}
\end{table*}

\section{Experiments}\label{sec:experiments}
\subsection{Datasets}
We evaluate the proposed method on two datasets: the 58-languages TED talk corpus~\citep{qi18naacl} and WMT datasets. 
For the TED corpus, we evaluate on two sets of languages with varying levels of language diversity following \citet{wang2020balancing}: (1) \emph{related} includes 4 LRLs (\texttt{aze}, \texttt{bel}, \texttt{glg}, \texttt{slk}) and their corresponding related HRL (\texttt{tur}, \texttt{rus}, \texttt{pos}, \texttt{ces}). (2) \emph{diverse} includes 8 languages with varying amount of data without considering linguistic similarities (\texttt{bos}, \texttt{mar}, \texttt{hin}, \texttt{mkd}, \texttt{ell}, \texttt{bul}, \texttt{fra}, \texttt{kor})\footnote{See \citet{wang2020balancing} for the interpretation of the language codes.}. 
Both of the related and diverse sets have around 760K sentences of training data.

For WMT, we consider 2 HRLs (German:\texttt{deu} and French:\texttt{fra}) and 2 LRLs (Tamil:\texttt{tam} and Turkish:\texttt{tur}). We subsample around 5M training sentences from the parallel corpus provided by the WMT shared task. Specifically, the training data of \texttt{deu}-\texttt{eng}, \texttt{fra}-\texttt{eng} is from WMT14, \texttt{tam}-\texttt{eng} is from WMT20 and \texttt{tur}-\texttt{eng} is from WMT18. We use the corresponding test and dev sets from each shared task for evaluation and validation.

We evaluate both \emph{en-to-any} (translate English to a target language) and \emph{any-to-en} (translate a source language into English) directions for all language sets. We provide dataset statistics in Appendix~\ref{app:datastats}.

\begin{table*}[h!]
\small
    \centering
    \begin{tabular}{l| cccc c| cccc c}
    \toprule
        \multirow{2}{*}{\textbf{Method}} & \multicolumn{5}{c|}{any$\rightarrow$en} & \multicolumn{5}{c}{en$\rightarrow$any}  \\
       & \textbf{deu} & \textbf{fra} & \textbf{tam} & \textbf{tur} & \textbf{Avg} & \textbf{deu} & \textbf{fra} & \textbf{tam} & \textbf{tur} & \textbf{Avg} \\
    \midrule
        FastDRO & 25.14 & 27.58 & 12.71 & 15.54 & 20.24 & 21.39 & 28.21 & 8.88 & 12.74 & 17.81 \\
        GDRO with PD & 26.72 & 29.13 & 15.78 & \underline{21.89} & 23.38 & 20.81 & 29.43 & \underline{10.29} & 15.52 & 19.01 \\
        CVaR-GDRO with PD & 28.62 & 30.70 & 15.94 & 20.61 & 23.97 & 22.81 & \underline{32.44} & 9.68 & 14.33 &  19.82 \\
        CVaR-GDRO with IBR & 29.14 & \underline{31.65} & \underline{16.31} & 20.98 & 24.52 & 22.34 & 31.97 & 10.15 & 14.82 & 19.82 \\
        
        $\chi^2$-GDRO with PD & 29.49 & 31.47 & 16.07 & 21.24 & 24.57 & \underline{23.10} & 32.30 & 9.87 & 14.70 &  19.99 \\
        ERM ($\tau$=5) & \underline{29.25} & 31.60 & \underline{16.31} & \underline{21.89} & \underline{24.76} & 22.67 & 32.36 & 10.04 & \textbf{16.09} & \underline{20.29} \\
        \ibr & \textbf{29.67} & \textbf{31.75} & \textbf{16.48} & \textbf{22.33} & \textbf{25.06} & \textbf{23.45} & \textbf{33.16} & \textbf{10.73} & \underline{15.53} & \textbf{20.72} \\
    \bottomrule
    \end{tabular}
    \caption{BLEU scores of different DRO objectives and algorithms---primal-dual (PD) and iterated best response (IBR)---on the WMT test sets.}
    \label{tab:wmt:dro}
    \vspace{-3mm}
\end{table*}
\begin{figure*}[t]
    \centering
      \includegraphics[width=0.85\textwidth]{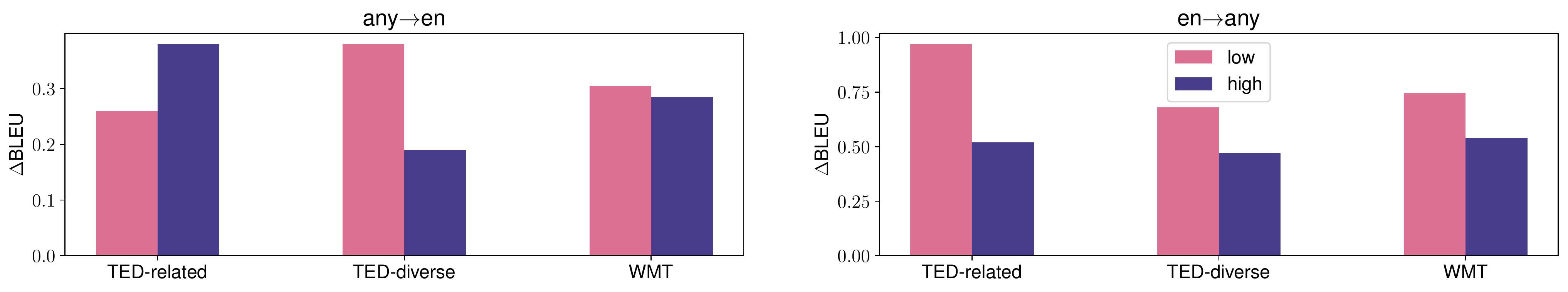}
      \vspace{-2mm}
    \caption{$\Delta$BLEU of low- and high-resource language groups for the three language sets. $\Delta$BLEU = difference of BLEU scores of \ibr and the best ERM model.}
    \label{fig:deltableu}
    \vspace{-1mm}
\end{figure*}

\begin{figure*}[t]
    \centering
      \includegraphics[width=0.95\textwidth]{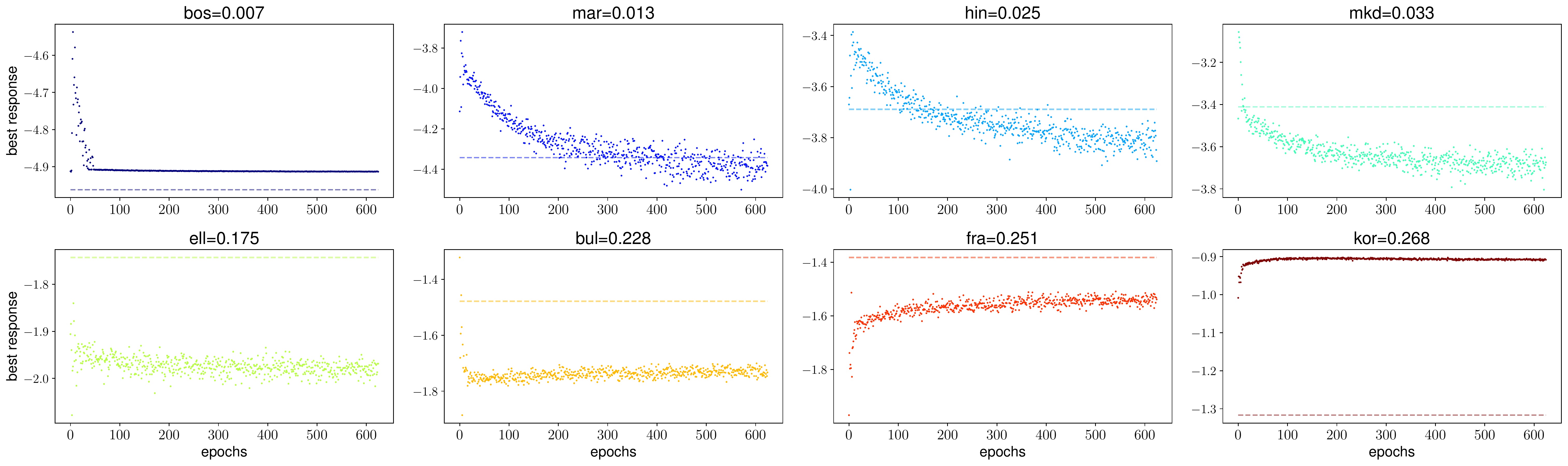}
      \vspace{-2mm}
    \caption{Best response $q$ (in log-scale) across epochs on the TED diverse dataset for the any$\rightarrow$en direction.}
    \label{fig:diverse:m2o:q}
    \vspace{-3mm}
\end{figure*}

\begin{figure*}[h]
    \centering
      \includegraphics[width=0.95\textwidth]{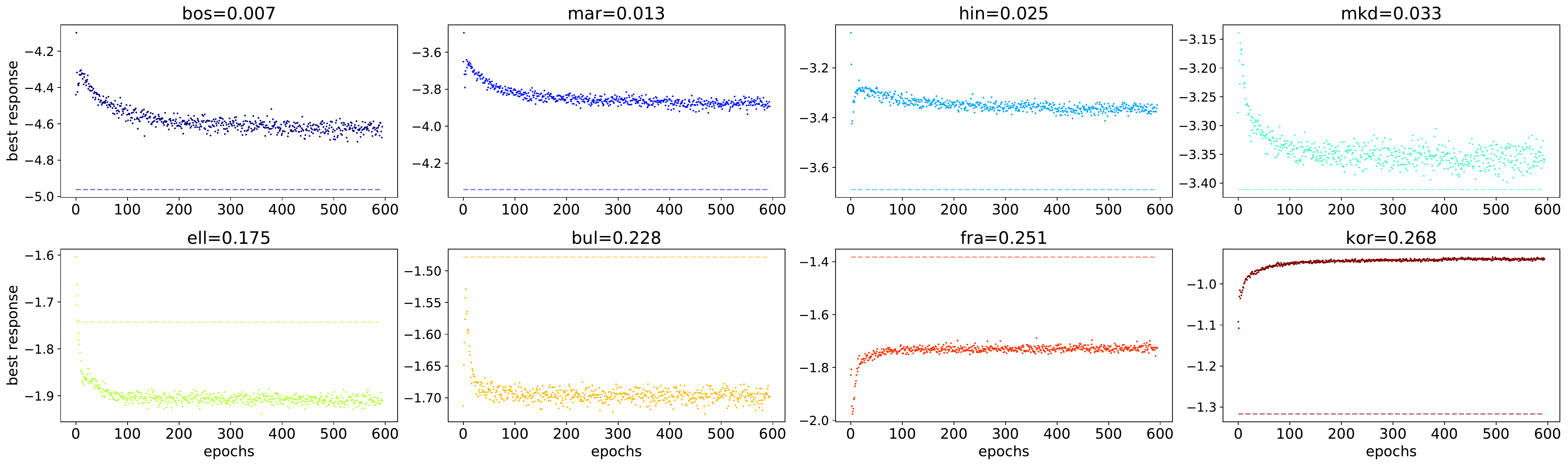}
      \vspace{-2mm}
    \caption{Best response $q$ (in log-scale) across epochs on the TED diverse dataset for the en$\rightarrow$any direction, the dashed line is the true data probability (in log-scale).}
    \label{fig:diverse:o2m:q}
    \vspace{-3mm}
\end{figure*}
\subsection{Experimental setup}
For the translation models, we adopt the encoder-decoder Transformer~\citep{vaswani2017attention} architecture with the implementation provided in fairseq~\citep{ott2019fairseq}. 
For both datasets, we use a Transformer-base architectures that also has 6 encoder and decoder layers with hidden dimension size being 512 and 8 attention heads.\footnote{We train for more steps with a larger batch size, which yields much better results than reported in~\citet{wang2020balancing}.}
The model is trained for 200K and 300K steps for TED and WMT respectively with the batch size of 65,536 tokens.
For both datasets, we learn the sentencepiece~\citep{kudo2018sentencepiece} vocabulary for the English and the combined corpus of other languages respectively. 
We use beam search with beam size 5 for decoding and report the SacreBLEU score~\citep{post-2018-call,Papineni:2002} on test sets for evaluation. 
For the TED and WMT datasets respectively, the constraint size $\rho$ for the chi-square ball is set to be 0.05 and 0.3, and
for the baseline losses we use the average token-level loss on each $D_i$ computed from the ERM model with $\tau=1$ and $\tau=100$---see \S\ref{sec:analysis} for more analyses of these choices. We provide additional pre-processing and training details in Appendix~\ref{app:train}.
\vspace{-2mm}
\paragraph{Baselines.}
We compare with (1) \textbf{temperature-based sampling} method described in \S\ref{sec:mmt} in three standard settings ($\tau=1/5/100$), where $\tau=100$ approximates uniform sampling over language pairs, and (2) \textbf{MultiDDS} described in \S\ref{sec:mmt}.
In addition, we also perform extensive empirical studies over different DRO uncertainty sets and optimization procedures in \S\ref{sec:analysis}.

\subsection{Main Results}
We present the BLEU scores of en$\rightarrow$any and any$\rightarrow$en translation directions on TED and WMT data in Tab.~\ref{tab:ted:full} and \ref{tab:wmt:full} respectively. 
First, for both TED and WMT datasets, \ibr outperforms all the other baseline methods in terms of average BLEU score over all language pairs.
By taking a closer look at the BLEU scores for each individual language pairs, \ibr improves over almost all the language pairs for both translation directions compared to ERM.
Secondly, as expected from temperature-based sampling methods, different values of $\tau$ achieve different trade-offs between the performances on HRLs and LRLs. As we explained in~\S\ref{sec:mmt}, large values of $\tau$ favor LRLs as this results in data being sampled with equal probability from each language pair while small values of $\tau$ approach ERM and will benefit HRLs. As a result, $\tau$ needs to be carefully tuned to achieve adequate performance on both HRLs and LRLs. 

Importantly, \ibr achieves a significantly better trade-off than the sampling method for any value of $\tau$. We show in Fig.~\ref{fig:deltableu} the quantitative improvements of \ibr over the best $\tau$ for various datasets and in both translation directions. Surprisingly, while the improvements are larger on LRLs in most cases, we consistently observe improvements on HRLs. This indicates that finding the right sampling distribution over language pairs facilitates cross-lingual transfer.
We further observe that \ibr achieves more significant improvements in the en$\rightarrow$any direction than in the any$\rightarrow$en direction. This further supports our hypothesis. 
Indeed, it is attested in previous work on MNMT~\citep{arivazhagan2019massively} that it is harder to decode to multiple languages than encode from multiple languages and as such, the en$\rightarrow$any direction is a significantly harder multi-task learning problem. As such, this is where optimal cross-lingual transfer would yield the larger gains, which is what we observe in practice.
We also note that our method incurs negligible computational overhead compared to ERM.
\section{Analysis}
\label{sec:analysis}
\paragraph{The importance of the sampling distribution.} 

An advantage of our method over temperature-based sampling methods is that it dynamically adjusts the training distribution as the model evolves and does not compute it solely as a function of amount of training data. Our hypothesis is that this is important to achieve good performance across language pairs and that different sampling distributions will be adequate at different stages of training. We empirically check our hypothesis by studying how the training distribution $q$ (the so-called best response) changes across training epochs. In Fig.~\ref{fig:diverse:m2o:q} and \ref{fig:diverse:o2m:q}, we plot the best response of \ibr 
across epochs on the TED-diverse dataset for both translation directions. In addition, we also plot the historical losses in Fig.~\ref{fig:ted:diverse:ema:loss} (Appendix). Our first observation is that the optimal $q$ noticeably evolves across epochs which further showcases the need for dynamically adjusting the sampling distribution. We make the following observations (i) \ibr demonstrates the desired behavior and, at the early stages of training, always down-weights HRLs and up-weights LRLs; (ii) somewhat counter-intuitively, there is no direct correlation between $\abs{D_i}$, the amount of data in language $i$ and the final value $\LL(\theta^{(t)};D_i)$. The latter further evidences the limitations of sampling distributions only based on the amounts of training data $\abs{D_i}$. Indeed, while \texttt{kor} is a HRL, it is typologically much farther from English so there is more inherent uncertainty in the task.
Consequently, it has larger losses and is consistently up-weighted throughout training. On the other hand, while \texttt{hin} is a LRL, it achieves low loss after being up-weighted during the early stages of training and is consequently down-weighted after that.

\vspace{-2mm}
\paragraph{Comparison to DRO variants.}
We demonstrate the benefits of \ibr over other DRO objectives by extensively evaluating a range of robust objectives and associated optimization algorithms. In terms of objective, we compare against (1) \textbf{Group DRO}~\cite{sagawa2019distributionally}, (2) \textbf{CVaR-Group DRO}~\cite{oren2019distributionally} and (3) \textbf{FastDRO}~\cite{LevyCaDuSi20}. In terms of algorithms, we experiment with primal-dual methods and our proposed iterated best response procedure which we both described in \S\ref{sec:opt:alg}. Note that in the case of Group DRO (i.e. $\mc{U} = \Delta^N$), iterated best response is not a sensible choice as it would result in each training epoch being spent on a single language pair. In the case of CVaR Group DRO, we follow the implementation of~\cite{oren2019distributionally}, which is a hybrid of the two optimization algorithms with a primal update on $\theta$ and a best response update on $q$. We compare the performance of these methods on the WMT dataset. For fairness, we baseline losses in the same way for all the DRO objectives. We first observe that, outside of \ibr\!, none of the DRO objectives are competitive with temperature-weighted ERM. We also observe that for both uncertainty sets, iterated best response convincingly outperforms the same objective trained with primal-dual. We finally note that, for a fixed optimization algorithm, $\cs$-Group DRO outperforms the CVaR objective on all but one language pairs. This validates both our choice of uncertainty set and of optimization procedure.

\begin{table}[t]
\scriptsize
    \centering
    \begin{tabular}{c|l|c}
    \toprule
    Dataset & Setting & Avg BLEU \\
    \midrule
    \parbox[t]{2mm}{\multirow{7}{*}{\rotatebox[origin=c]{90}{TED}}} & (a) ERM, $\tau=1$ 	& 21.46 \\
    & (b) ERM, $\tau=100$ & 20.41 \\
    & (c) Ours, $\rho=0.05$, w/o BL & 22.08 \\
    & (d) Ours, $\rho=0.1$, w/o BL & 21.75 \\
    & (e) Ours, $\rho=0.05$, BL: $\tau=1$ & 22.21 \\
    & (f) Ours, $\rho=0.1$, BL: $\tau=1$ & 22.13 \\
    & (g) Ours, $\rho=0.05$, BL: $\tau=100$ & 21.37 \\
    \midrule
    \parbox[t]{2mm}{\multirow{2}{*}{\rotatebox[origin=c]{90}{WMT}}} & (h) Ours, $\rho=0.1$, BL: $\tau=1$ & 20.34 \\
    & (i) Ours, $\rho=0.1$, BL: $\tau=100$ & 20.62 \\
    \bottomrule
    \end{tabular}
    \caption{Average BLEU on the test sets of en$\rightarrow$any direction, BL is short for baseline loss.}
    \label{tab:abl:baseline}
    \vspace{-5mm}
\end{table}
\vspace{-2mm}
\paragraph{The effects of baselined losses.} We study the effect of the choice of baseline on the performance across languages. In Tab.~\ref{tab:abl:baseline}, we empirically evaluate different baseline choices and uncertainty sizes $\rho$. We observe that in the TED dataset, baseline-ing with $\LL(\what{\theta}^{\mathsf{ERM}};D_i)$ performs significantly better than baseline-ing with $\LL(\what{\theta}^{\tau=100};D_i)$ ((e) versus (g) while it is reversed for WMT. We explain this by observing that the LRLs in TED consist of very small amounts of data (on the order of a few thousands) and using $\tau=100$ results in a severe oversampling of LRLs, which the model then fits perfectly. As a result, recall the intuition that the baseline sets a lower bound on the performance we wish to achieve but because of the small training data and overfitting, the model disproportionately up-weighs the LRLs, which harms overall performance. This does not occur in WMT and uniform sampling across language pairs sets a good target performance for DRO methods. Finally, we see that with the right baseline loss, our method is more robust to different choices of $\rho$ (e.g., comparing (c) and (d) versus (e) and (f)). 
We consistently observe this for other translation directions and datasets.



%% file: primal_dual.tex
\section{Best response}\label{app:best-response}

In this section, we describe the bisection procedure we use to solve the best response and update $q$ shown in~\eqref{eq:update-q}. This derivation generically exists in the literature (e.g. Appendix A.1.2 in~\cite{LevyCaDuSi20}) but we specialize it to the $\cs$-ball centered at $\ptrain$ and include it here for completeness.

For $v\in\R^m$, the optimization problem we wish to solve is
\begin{equation}\label{eq:best-response-cs}
\begin{split}
    \maximize_{q\in \Delta^m} & \qquad q^\top v \\
    \subjectto & \qquad \cs(q, \ptrain) \le \rho,
\end{split}
\end{equation}

Let us consider the Lagrangian of this problem
\begin{equation*}
    \Lambda(q;\nu, \eta) \defeq q^\top v - \eta(\ones^\top q - 1) - \lambda(\cs(q, \ptrain) - \rho)
\end{equation*}
Maximizing over $q$ yields that the solution as a function of $\eta$ is $q^\star(\eta)_i \propto \ptrain_i (v_i - \eta)_+$, where $(u)_+ \defeq \max\crl{0, u}$. Since the objective is linear, it holds that the maximum is attained on the extreme points of the constraint sets. Consequently, we need to find $\eta^\star$ such that $\cs(q^\star(\eta^\star), \ptrain) = \rho$. This is a simple root finding procedure that we solve to accuracy $\epsilon$ in $\log (1/\epsilon)$ steps with a bisection. Given that each evaluation of $q^\star(\eta)$ requires $O(m)$ operations, the runtime of the algorithm is $O(m\log(1/\epsilon))$.

\section{Primal-dual methods}\label{app:primal-dual}

Primal-dual algorithms~\cite{Nemirovski04, NemirovskiJuLaSh09} are the methods of choice to efficiently solve min-max problems.

\subsection{The primal-dual algorithm}

Let us assume that $\mc{X} \subset \R^d$ and $\mc{Y} \subset \R^p$ are closed, bounded convex sets and let $F:\mc{X}\times\mc{Y}\to \R$ be a function such that $F(x, \cdot)$ is concave for all $x\in\mc{X}$ and $F(\cdot, y)$ is convex for all $y\in\mc{Y}$. We wish to solve the following min-max problem
\begin{equation}\label{eq:minmax}
    \min_{x\in\mc{X}} \max_{y\in\mc{Y}} F(x, y).
\end{equation}

This is a well-studied problem of optimization and there the literature provides optimal algorithms for many choices of $F$, $\mc{X}$ and $\mc{Y}$. A standard approach are the so-called \emph{primal-dual methods}, where one keeps a pair of iterates $(x_t, y_t)$ performs a gradient descent (resp. ascent) step on $x_t$ (resp. $y_t$). In all generality, these updates are often mirror descent updates to properly exploit the geometric structures of $\mc{X}, \mc{Y}$ and $F$. Let $h^\msf{x}:\R^d\to\R$ (resp. $h^\msf{y}:\R^p\to\R$) be a $1$-strongly-convex function w.r.t. a given norm $\norms{\cdot}_\msf{x}$ (resp. $\norms{\cdot}_\msf{y}$). We denote $D_{h^\msf{x}}$ and $D_{h^\msf{y}}$ their associated Bregman divergences. The primal-dual algorithms initialize $x_0 \in \mc{X}, y_0 \in\mc{Y}$ and for a stepsize $\eta > 0$, iterates
\begin{equation*}
    \begin{split}
        x_{t+1} & = \argmin_{x\in\mc{X}} \crl*{
        g_{\msf{x}, t}^\top x + \frac{1}{\eta}D_{h^\msf{x}}(x, x_t)
        } \\
        y_{t+1} & = \argmax_{y\in\mc{Y}} \crl*{
        g_{\msf{y}, t}^\top y - \frac{1}{\eta}D_{h^\msf{y}}(y, y_t),
        },
    \end{split}
\end{equation*}
where $g_{\msf{x}, t} \in \partial_x F(x_t, y_t)$ and $g_{\msf{y}, t} \in \partial_y F(x_t, y_t)$. After $T$ steps, return $(\bar{x}_T, \bar{y}_T)$ with $\bar{z}_T \defeq 1/T\sum_{t\le T}z_t$. Assuming $F$ is appropriately Lipschitz and that $\mc{X}$ and $\mc{Y}$ are bounded, one finds an $\epsilon$-approximate saddle point in $O(\epsilon^{-2})$ steps. Importantly, these guarantee still holds even when only having \emph{stochastic unbiased estimates} of $g_\msf{x}$ and $g_\msf{y}$~\cite{NemirovskiJuLaSh09} which is essential in large-scale settings. 

\subsection{Primal-dual algorithms for Group DRO}

The problem~\eqref{eq:gen-gdro} can be cast as an instance of~\eqref{eq:minmax} by setting $\mc{Y} = \mc{U}$ and $F(\theta, q) \defeq \sum_{i\le N}q_i\LL(\theta;D_i)$. Note that the problem is unconstrained in $\theta$ but it is generally not an issue in practice. The gradient in $\theta$ and $q$ are
\begin{equation*}
    \begin{split}
        \nabla_\theta F(\theta, q) = \sum_{i\le N}q_i \nabla_\theta \LL(\theta;D_i) \\
        \brk*{\nabla_q F(\theta, q)}_i = \LL(\theta;D_i).
    \end{split}
\end{equation*}

\subsubsection{Obtaining stochastic gradients} To run primal-dual, we require stochastic gradient estimates $\tilde{g}_{\theta}$ and $\tilde{g}_{q}$. First of all, note that an unbiased estimate of $\nabla_\theta \LL(\theta;D_i)$ is just $\nabla_\theta \ell(\vx, \vy;\theta)$ where $(\vx, \vy)$ is sampled at random from $D_i$. To obtain stochastic gradient estimates, we have two choices: sampling from $q$ or sampling from an arbitrary $p_0 \in \Delta^N$ and importance-weighting. 

\paragraph{Sampling from $q$} Let $B$ be the mini-batch size and $\prn{I_1, \ldots, I_B}$ be $B$ indices sampled from $q$ and $(\vx_{I_j}, \vy_{I_j})$ be randomly sampled examples from $D_{I_j}$. In this case,
\begin{equation*}
\begin{split}
    \tilde{g}_\theta & = \frac{1}{B}\sum_{j\le B} \nabla_\theta \ell(\vx_{I_j}, \vy_{I_j};\theta) \\
    \brk{\tilde{g}_q}_i & = \frac{1}{B}\sum_{j\le B} \frac{1}{q_{I_j}}\ell(\vx_{I_j}, \vy_{I_j};\theta),
    \end{split}
\end{equation*}
are clearly unbiased stochastic gradient estimates for $(\nabla_\theta F(\theta, q), \nabla_q F(\theta, q))$.

\paragraph{Sampling from an arbitrary $p_0 \in \Delta^N$} Assume $p_0$ is in $\Delta^n$ and that $p_{0, i} > 0$ for $i\in\brk{N}$. As previously, assume we sample a mini-batch of indices $\prn{J_1, \ldots, J_B}$ from $p_0$ and that $(\vx_{J_j}, \vx_{J_j})$ is a randomly sampled example from $D_{J_j}$. We can obtain stochastic gradient estimates by importance-weighting (IW)
\begin{equation*}
\begin{split}
    \tilde{g}^{\msf{IW}}_\theta & = \frac{1}{B}\sum_{j\le B} \frac{q_{J_j}}{p_{0, J_j}}\nabla_\theta \ell(\vx_{J_j}, \vy_{J_j};\theta) \\
    \brk{\tilde{g}^{\msf{IW}}_q}_i & = \frac{1}{B}\sum_{j\le B} \frac{1}{p_{0, I_j}}\ell(\vx_{I_j}, \vy_{I_j};\theta),
    \end{split}
\end{equation*}

In terms of implementation, it is often impractical to change the distribution of batches at each iteration of the optimizer. 

\subsubsection{Choice of Bregman divergence}

\paragraph{$\theta$-update.} The update in $\theta$ is unconstrained and it is standard to optimize the objective with Stochastic Gradient Descent so as a result, we pick $h^\theta(\theta) = \frac{1}{2}\norm{\theta}_2^2$. This results in the standard SGD update
\begin{equation*}
    \theta_{t+1} = \theta_t - \eta \tilde{g}_\theta,
\end{equation*}
where $\tilde{g}_\theta$ is an unbiased stochastic gradient estimate that we compute following the previous section.

\paragraph{$q$-update for $\mc{U} = \Delta^N$.} In the case where $\mc{U}$ is the full simplex, it is standard to choose $h^\msf{q}(q) = \sum_{i\le N}q_i \log q_i$ (the negative Shannon entropy). This choice yields the familiar Exponentiated Gradient update
\begin{equation*}
    q_{t+1, i} \propto q_{t, i}\exp(\eta \brk{\tilde{g}_q}_i)
\end{equation*}

\paragraph{$q$-update for $\mc{U} = \crl{q\in\Delta^N: \cs(q, \ptrain) \le \rho}$.} As the $\cs$-uncertainty set is essentially the intersection between a (weighted)-norm-2 ball and the simplex, it is a natural choice to pick $h^\msf{q}(q) = \frac{1}{2}\norm{q}_2^2$. This leads to the following update
\begin{equation*}
    q_{t+1} = \argmin_{q\in\Delta^N:\cs{q, \ptrain}\le \rho}\norm*{(q_t + \eta \tilde{g}_q) - q}_2^2,
\end{equation*}
or in other words, the projection of the gradient ascent step $(q_t + \eta \tilde{g}_q)$ onto the $\cs$-ball. We explain in Appendix~\ref{app:projection} how to efficiently compute this projection for arbitrary $\ptrain$.

\paragraph{$q$-update for $\mc{U} = \mc{U}_\alpha$.} We follow the implementation of~\citet{oren2019distributionally} which runs a hybrid between primal-dual methods and best response and thus we do not need to explicit include the projection onto the CVaR uncertainty set. We discuss the option here for completeness. For the CVaR uncertainty set, it is standard to also use the negative Shannon entropy $h^\msf{q} = \sum_i q_i \log q_i$. We refer to Appendix~F.6.2 of~\cite{LevyCaDuSi20} and their provided code for more details on this projection.

\subsection{Projecting on the $\cs$-ball}\label{app:projection}

To run the primal-dual algorithm, the gradient update on $q$ requires projecting on the $\cs$-ball centered at $\ptrain$. For $v\in\R^m$, we wish to solve the following
\begin{equation}\label{eq:proj-cs}
\begin{split}
    \minimize_{q\in \Delta^m} & \qquad \norm{q - v}_2^2 \\
    \subjectto & \qquad \cs(q, \ptrain) \le \rho,
\end{split}
\end{equation}
where $\Delta^m$ is the $m$-dimensional simplex and $\cs(q, \ptrain) \defeq \tfrac{1}{2}\sum_{i\le m} p_i (q_i / p_i - 1)^2$, the $f$-divergence~\cite{Csiszar67} corresponding to $t\mapsto \tfrac{1}{2}(t-1)^2$. 

While this projection is standard in the literature, it is often derived in the case of $\ptrain = \ones / m$ (see e.g.~\cite{NamkoongDu16}). We show here how to efficiently do the projection for arbitrary $\ptrain \in \Delta^m$.

First, for $\lambda \ge 0, \eta\in\R$, the Lagrangian of~\eqref{eq:proj-cs} is
\begin{equation*}
\begin{split}
    \Lambda(q, \lambda, \eta) \defeq & \frac{1}{2}\norm{q - v}_2^2 \\
    & + \lambda (\cs(q, \ptrain) - \rho) + \eta(q^\top \ones - 1).
    \end{split}
\end{equation*}
Taking the partial dual $g(\lambda, \eta) \defeq \inf_{q \succeq 0} \Lambda(q, \lambda, \eta)$ yields
\begin{equation*}
\begin{split}
    g(\lambda, \eta) = -\inf_{\lambda \ge 0, \eta \in \R} & \frac{1}{2} \sum_{i\le m} \frac{p_i}{p_i + \lambda}(v_i - \eta)_+ \\ 
    & + \frac{\lambda}{2}(1+2\rho) + \eta
    \end{split}
\end{equation*}
In contrast to the uniform case (i.e.\ $\ptrain = \ones/m$), one cannot derive the optimal dual variable $\lambda^*$ in closed-form and we have to solve for both dual variables. Finding an $\epsilon$-accurate solution takes order $m\log(1/\epsilon)$ time using cutting plane-type methods when the dimension is $O(1)$~\cite{Bubeck15}. In the large-scale applications we consider, this is negligible in comparison to computing the gradient of the loss with respect to the network parameters and thus the primal-dual algorithm incurs (almost) no additional computational overhead.

In practice, we implement this with two nested bisections; while this adds an additional $\log(1/\epsilon)$ factor, this is significantly more convenient to implement.

%% file: appendix.tex
\section{Data Statistics}
\label{app:datastats}
\begin{table}[h]
    \centering
    \begin{tabular}{lc|lc}
    \toprule
        related & \#sents & diverse & \#sents \\
    \midrule
        bel	& 4,509 & bos & 5,664\\
        aze	& 5,946 & mar & 9,840 \\
        glg	& 10,017 & hin & 18,798\\
        slk	& 61,470 & mkd & 25,335\\
        cse	& 103,093 & ell & 134,327\\
        tur	& 182,470 & fra & 192,304\\
        por	& 184,755 & bul & 174,444\\
        rus	& 208,458 & kor & 205,640\\
    \bottomrule
    \end{tabular}
    \caption{Number of training sentences in the TED related and diverse sets respectively.}
    \label{tab:ted:stats}
\end{table}
We present the number of training sentences in the TED datasets in Tab.~\ref{tab:ted:stats}.
The number of training sentences in the WMT dataset is 2.5M (\texttt{deu}), 1.8M (\texttt{fra}), 512,608 (\texttt{tam}) and 195,762 (\texttt{tur}).

\begin{figure*}%
    \centering
    \subfloat{{\includegraphics[width=0.45\textwidth]{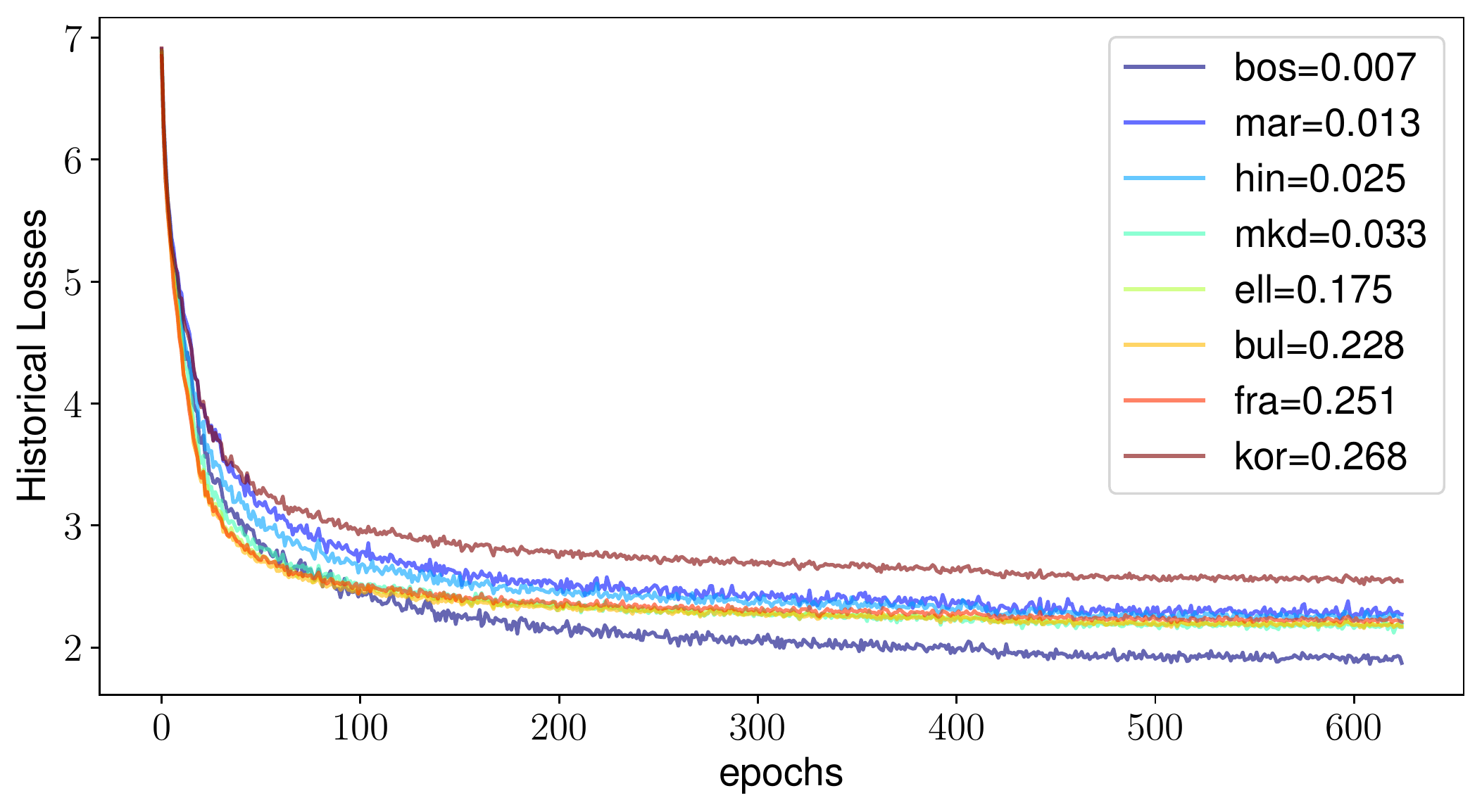} }}
    \qquad
    \subfloat{{\includegraphics[width=0.45\textwidth]{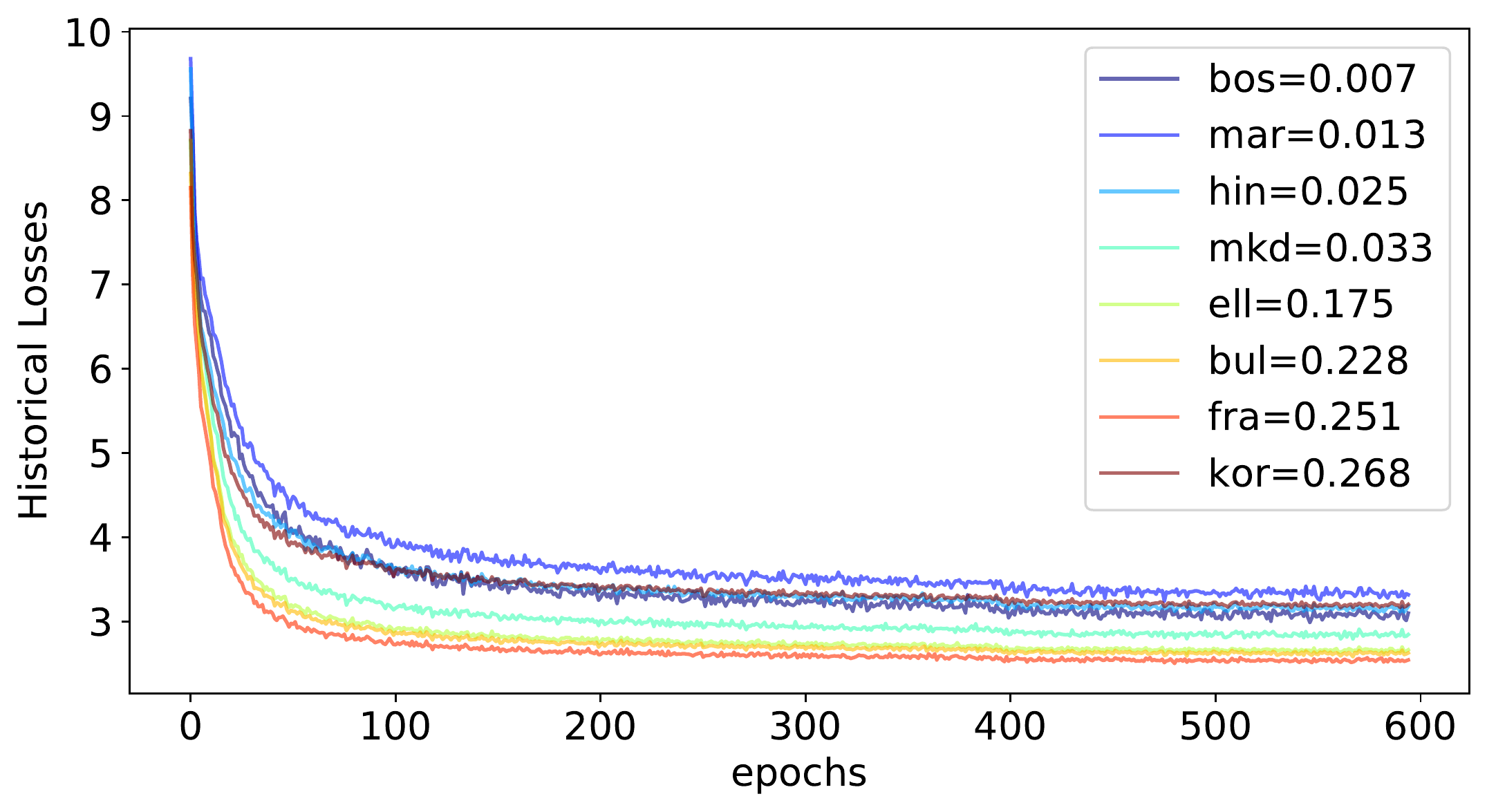} }}
    \vspace{-3mm}
    \caption{The historical (EMA) training losses on the TED-diverse dataset (left: any$\rightarrow$en, right: en$\rightarrow$any).}
    \label{fig:ted:diverse:ema:loss}
    \vspace{-3mm}
\end{figure*}

\begin{figure*}%
    \centering
    \subfloat{{\includegraphics[width=0.45\textwidth]{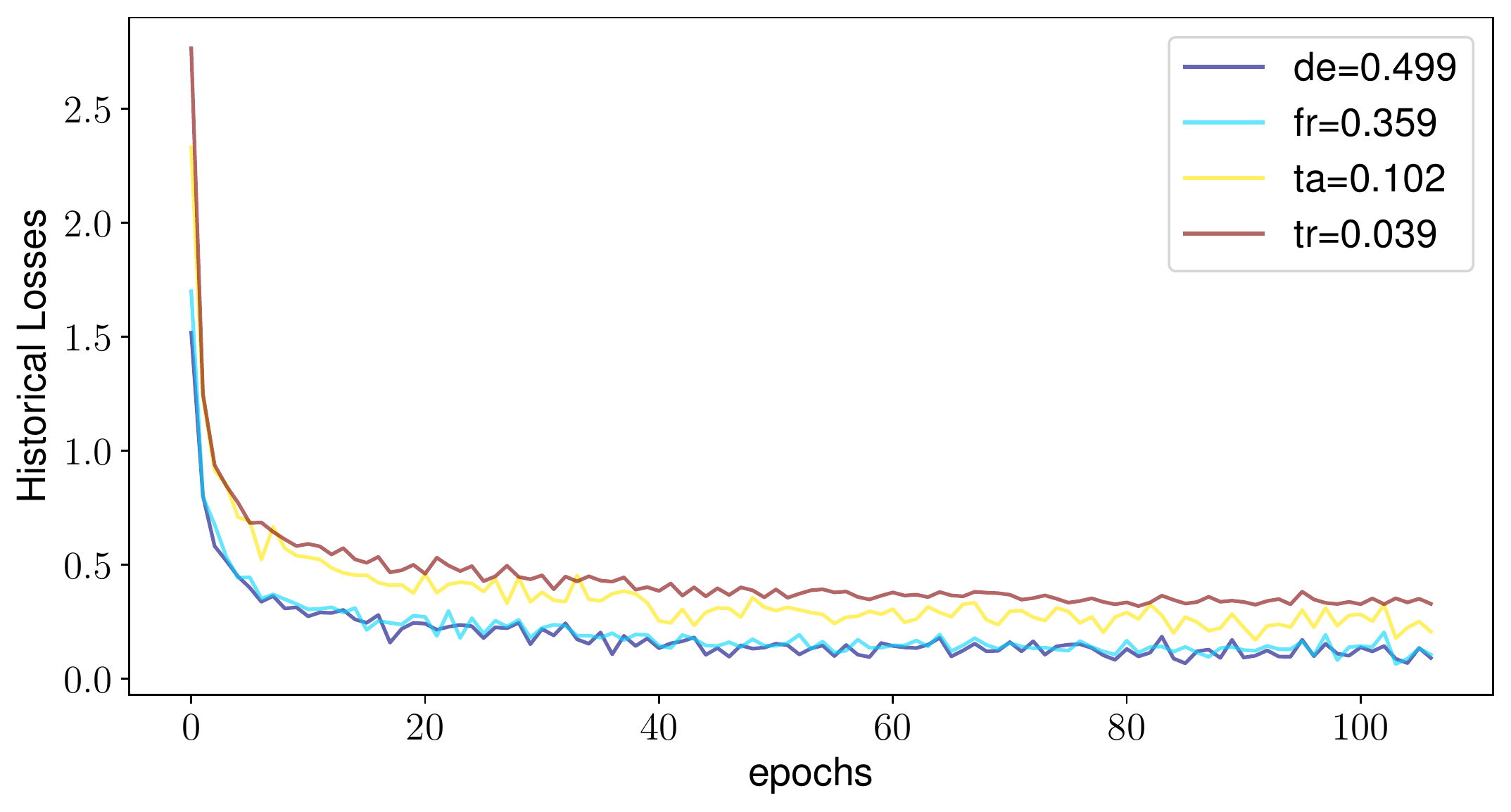} }}
    \qquad
    \subfloat{{\includegraphics[width=0.45\textwidth]{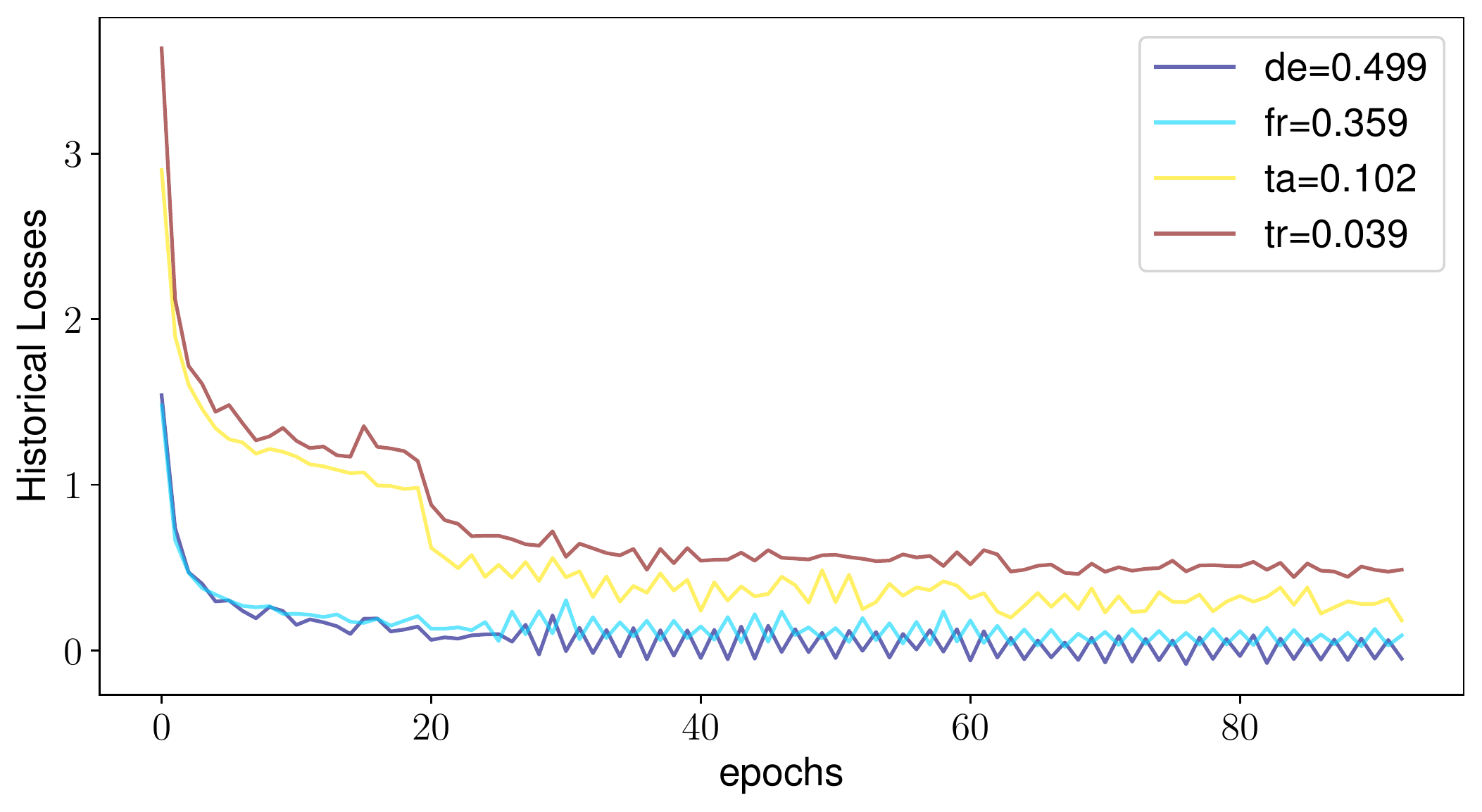} }}
    \vspace{-3mm}
    \caption{The historical (EMA) training losses on the WMT dataset (left: any$\rightarrow$en, right: en$\rightarrow$any).}
    \label{fig:wmt:ema:loss}
    \vspace{-3mm}
\end{figure*}

\begin{figure*}[h!]
    \centering
      \includegraphics[width=0.95\textwidth]{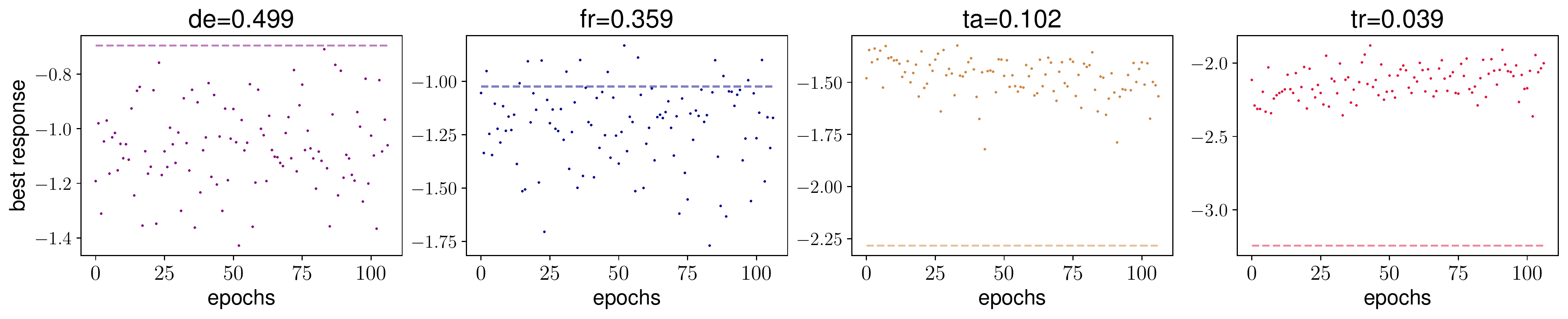}
      \vspace{-2mm}
    \caption{Best response $q$ (in log-scale) across epochs on the WMT dataset for the any$\rightarrow$en direction, the dashed line is the true data probability (in log-scale).}
    \label{fig:wmt:o2m:q}
    \vspace{-3mm}
\end{figure*}
\begin{figure*}[h!]
    \centering
      \includegraphics[width=0.95\textwidth]{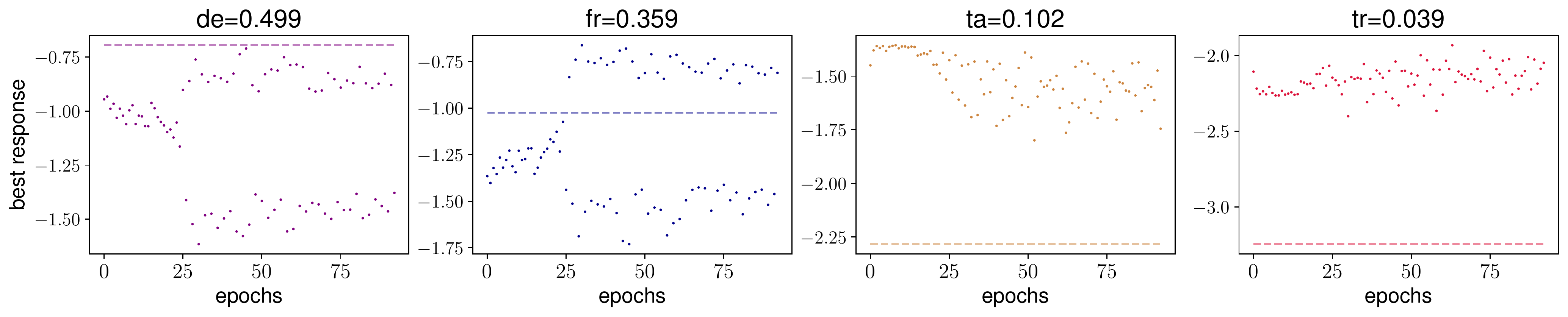}
      \vspace{-2mm}
    \caption{Best response $q$ (in log-scale) across epochs on the TED diverse dataset for the en$\rightarrow$any direction, the dashed line is the true data probability (in log-scale).}
    \label{fig:wmt:m2o:q}
    \vspace{-3mm}
\end{figure*}

\section{Preprocessing and Training Details}
\label{app:train}
We describe the preprocessing and training details in this section. 
\paragraph{Preprocessing} 
We download WMT datasets from the WMT official websites\footnote{e.g. \url{http://www.statmt.org/wmt18/}}.
The TED data is downloaded from the link\footnote{\url{https://github.com/cindyxinyiwang/multiDDS}.} provided by \citet{wang2020balancing}.
We learn sentencepiece~\citep{kudo2018sentencepiece} for English and other languages respectively. Specifically, we combine all the English sentences from 
individual parallel datasets. Following \citet{arivazhagan2019massively}, we also resample the sentences of other languages based on the temperature-based distribution ($\tau=5$) to learn the vocabulary. For TED, the vocabulary sizes for English and other languages are set to be 10K and 30K respectively. For WMT, the vocabulary sizes for English and other languages are set to be 24K and 34K respectively.
We also remove sentences in the training data that are longer than 250 tokens after bpe.
For validation set, to ensure the validation loss are fairly comparable across different languages, we cap the number of the sentences in the validation set to be the same for each language (800 for TED and 1,500 for WMT).
\begin{table}[h]
    \centering
    \begin{tabular}{ccc}
    \toprule
    Models & \textit{small} & \textit{base} \\
    \midrule
    $d_\textrm{model}$ & 512 & 512  \\
    $d_\textrm{hidden}$ & 1024 & 2048 \\
    $n_\textrm{layers}$ & 6 & 6  \\
    $n_\textrm{heads}$ & 4 & 8 \\
    $p_\textrm{dropout}$ & 0.3 & 0.3 \\
    \bottomrule
    \end{tabular}
    \caption{Basic hyper-parameters of Transformer.}
    \label{tab:at}
\end{table}

\paragraph{Training details} 
The hyperparameters of Transformer models are described in Tab.~\ref{tab:at}.
For all experiments, we adopt the Adam optimizer~\citep{kingma2014adam} using $\beta_1=0.9, \beta_2=0.98, \epsilon=1e-8$. We set the label smoothing as $0.1$\footnote{Note that we also use the label smoothing loss as the baseline loss from trained ERM models}.
For ERM models, we use the default learning rate scheduling. 
The learning rate is scheduled using \texttt{inverse\_sqrt} with a maximum learning rate of $5e-4$ and $2e-4$ (following~\citep{wang2020balancing}) for WMT and TED respectively. The warmup steps is set to be 4000 for both datasets. 
For \ibr, because the training distribution is dynamically changing across epochs, we use step learning rate decay to retain a higher learning rate during training. 
Specifically, we use the same maximum learning rate and warm up steps as ERM, but decay the learning rate by half every 100K training steps for both datasets.
Recall that we use exponential moving average to compute the historical loss values, and we set the hyperparameter $\lambda$ to be 0.1 throughout.
We train on 4 and 8 V100 GPUs for TED and WMT experiments respectively. 
The training time for one experiment takes around 1 - 2 days.

%% file: emnlp2021.bbl
\begin{thebibliography}{38}
\expandafter\ifx\csname natexlab\endcsname\relax\def\natexlab#1{#1}\fi

\bibitem[{Aharoni et~al.(2019)Aharoni, Johnson, and
  Firat}]{aharoni2019massively}
Roee Aharoni, Melvin Johnson, and Orhan Firat. 2019.
\newblock Massively multilingual neural machine translation.
\newblock In \emph{Proceedings of the 2019 Conference of the North American
  Chapter of the Association for Computational Linguistics: Human Language
  Technologies, Volume 1 (Long and Short Papers)}, pages 3874--3884.

\bibitem[{Arivazhagan et~al.(2019)Arivazhagan, Bapna, Firat, Lepikhin, Johnson,
  Krikun, Chen, Cao, Foster, Cherry et~al.}]{arivazhagan2019massively}
Naveen Arivazhagan, Ankur Bapna, Orhan Firat, Dmitry Lepikhin, Melvin Johnson,
  Maxim Krikun, Mia~Xu Chen, Yuan Cao, George Foster, Colin Cherry, et~al.
  2019.
\newblock Massively multilingual neural machine translation in the wild:
  Findings and challenges.
\newblock \emph{arXiv preprint arXiv:1907.05019}.

\bibitem[{Ben-Tal et~al.(2013{\natexlab{a}})Ben-Tal, Den~Hertog, De~Waegenaere,
  Melenberg, and Rennen}]{ben2013robust}
Aharon Ben-Tal, Dick Den~Hertog, Anja De~Waegenaere, Bertrand Melenberg, and
  Gijs Rennen. 2013{\natexlab{a}}.
\newblock Robust solutions of optimization problems affected by uncertain
  probabilities.
\newblock \emph{Management Science}, 59(2):341--357.

\bibitem[{Ben-Tal et~al.(2013{\natexlab{b}})Ben-Tal, den Hertog, Waegenaere,
  Melenberg, and Rennen}]{Ben-TalHeWaMeRe13}
Aharon Ben-Tal, Dick den Hertog, Anja~De Waegenaere, Bertrand Melenberg, and
  Gijs Rennen. 2013{\natexlab{b}}.
\newblock Robust solutions of optimization problems affected by uncertain
  probabilities.
\newblock \emph{Management Science}, 59(2):341--357.

\bibitem[{Bertsimas et~al.(2018)Bertsimas, Gupta, and Kallus}]{BertsimasGuKa18}
Dimitris Bertsimas, Vishal Gupta, and Nathan Kallus. 2018.
\newblock Data-driven robust optimization.
\newblock \emph{Mathematical Programming, Series A}, 167(2):235--292.

\bibitem[{Boyd and Vandenberghe(2004)}]{BoydVa04}
Stephen Boyd and Lieven Vandenberghe. 2004.
\newblock \emph{Convex Optimization}.
\newblock Cambridge University Press.

\bibitem[{Bubeck(2015)}]{Bubeck15}
S{\'e}bastien Bubeck. 2015.
\newblock Convex optimization: Algorithms and complexity.
\newblock \emph{Foundations and Trends in Machine Learning}, 8(3-4):231--357.

\bibitem[{Conneau et~al.(2020)Conneau, Khandelwal, Goyal, Chaudhary, Wenzek,
  Guzm{\'a}n, Grave, Ott, Zettlemoyer, and Stoyanov}]{conneau2020unsupervised}
Alexis Conneau, Kartikay Khandelwal, Naman Goyal, Vishrav Chaudhary, Guillaume
  Wenzek, Francisco Guzm{\'a}n, {\'E}douard Grave, Myle Ott, Luke Zettlemoyer,
  and Veselin Stoyanov. 2020.
\newblock Unsupervised cross-lingual representation learning at scale.
\newblock In \emph{Proceedings of the 58th Annual Meeting of the Association
  for Computational Linguistics}, pages 8440--8451.

\bibitem[{Csisz\'ar(1967)}]{Csiszar67}
Imre Csisz\'ar. 1967.
\newblock Information-type measures of difference of probability distributions
  and indirect observation.
\newblock \emph{Studia Scientifica Mathematica Hungary}, 2:299--318.

\bibitem[{Delage and Ye(2010)}]{DelageYe10}
Erick Delage and Yinyu Ye. 2010.
\newblock Distributionally robust optimization under moment uncertainty with
  application to data-driven problems.
\newblock \emph{Operations Research}, 58(3):595--612.

\bibitem[{Devlin et~al.(2019)Devlin, Chang, Lee, and
  Toutanova}]{kenton2019bert}
Jacob Devlin, Ming-Wei Chang, Kenton Lee, and Kristina Toutanova. 2019.
\newblock Bert: Pre-training of deep bidirectional transformers for language
  understanding.
\newblock In \emph{Proceedings of NAACL-HLT}, pages 4171--4186.

\bibitem[{Duchi et~al.(2016)Duchi, Glynn, and Namkoong}]{duchi2016statistics}
John Duchi, Peter Glynn, and Hongseok Namkoong. 2016.
\newblock Statistics of robust optimization: A generalized empirical likelihood
  approach.
\newblock \emph{arXiv preprint arXiv:1610.03425}.

\bibitem[{Duchi and Namkoong(2019)}]{DuchiNa19}
John~C. Duchi and Hongseok Namkoong. 2019.
\newblock Variance-based regularization with convex objectives.
\newblock \emph{Journal of Machine Learning Research}, 20(68):1--55.

\bibitem[{Duchi and Namkoong(2020)}]{DuchiNa20}
John~C. Duchi and Hongseok Namkoong. 2020.
\newblock \href {https://arXiv.org/abs/1810.08750} {Learning models with
  uniform performance via distributionally robust optimization}.
\newblock \emph{Annals of Statistics}, to appear.

\bibitem[{Firat et~al.(2016)Firat, Cho, and Bengio}]{firat2016multi}
Orhan Firat, Kyunghyun Cho, and Yoshua Bengio. 2016.
\newblock Multi-way, multilingual neural machine translation with a shared
  attention mechanism.
\newblock In \emph{Proceedings of the 2016 Conference of the North American
  Chapter of the Association for Computational Linguistics: Human Language
  Technologies}, pages 866--875.

\bibitem[{Ha et~al.(2016)Ha, Niehues, and Waibel}]{ha2016toward}
Thanh-Le Ha, Jan Niehues, and Alexander Waibel. 2016.
\newblock Toward multilingual neural machine translation with universal encoder
  and decoder.
\newblock \emph{arXiv preprint arXiv:1611.04798}.

\bibitem[{Hashimoto et~al.(2018)Hashimoto, Srivastava, Namkoong, and
  Liang}]{HashimotoSrNaLi18}
Tatsunori Hashimoto, Megha Srivastava, Hongseok Namkoong, and Percy Liang.
  2018.
\newblock Fairness without demographics in repeated loss minimization.
\newblock In \emph{Proceedings of the 35th International Conference on Machine
  Learning}.

\bibitem[{Johnson et~al.(2017)Johnson, Schuster, Le, Krikun, Wu, Chen, Thorat,
  Vi{\'e}gas, Wattenberg, Corrado et~al.}]{johnson2017google}
Melvin Johnson, Mike Schuster, Quoc Le, Maxim Krikun, Yonghui Wu, Zhifeng Chen,
  Nikhil Thorat, Fernanda Vi{\'e}gas, Martin Wattenberg, Greg Corrado, et~al.
  2017.
\newblock Google’s multilingual neural machine translation system: Enabling
  zero-shot translation.
\newblock \emph{Transactions of the Association for Computational Linguistics},
  5:339--351.

\bibitem[{Kingma and Ba(2014)}]{kingma2014adam}
Diederik Kingma and Jimmy Ba. 2014.
\newblock Adam: A method for stochastic optimization.
\newblock \emph{arXiv preprint arXiv:1412.6980}.

\bibitem[{Kudo and Richardson(2018)}]{kudo2018sentencepiece}
Taku Kudo and John Richardson. 2018.
\newblock Sentencepiece: A simple and language independent subword tokenizer
  and detokenizer for neural text processing.
\newblock In \emph{Proceedings of the 2018 Conference on Empirical Methods in
  Natural Language Processing: System Demonstrations}, pages 66--71.

\bibitem[{Levy et~al.(2020)Levy, Carmon, Duchi, and Sidford}]{LevyCaDuSi20}
Daniel Levy, Yair Carmon, John~C. Duchi, and Aaron Sidford. 2020.
\newblock \href {https://arxiv.org/abs/2010.05893} {Large-scale methods for
  distributionally robust optimization}.
\newblock In \emph{Advances in Neural Information Processing Systems 33}.

\bibitem[{Namkoong and Duchi(2016)}]{NamkoongDu16}
Hongseok Namkoong and John~C. Duchi. 2016.
\newblock Stochastic gradient methods for distributionally robust optimization
  with $f$-divergences.
\newblock In \emph{Advances in Neural Information Processing Systems 29}.

\bibitem[{Nemirovski et~al.(2009)Nemirovski, Juditsky, Lan, and
  Shapiro}]{NemirovskiJuLaSh09}
A.~Nemirovski, A.~Juditsky, G.~Lan, and A.~Shapiro. 2009.
\newblock Robust stochastic approximation approach to stochastic programming.
\newblock \emph{SIAM Journal on Optimization}, 19(4):1574--1609.

\bibitem[{Nemirovski(2004)}]{Nemirovski04}
Arkadi Nemirovski. 2004.
\newblock Prox-method with rate of convergence {$O(1/t)$} for variational
  inequalities with {L}ipschitz continuous monotone operators and smooth
  convex-concave saddle point problems.
\newblock \emph{SIAM Journal on Optimization}, 15(1):229--251.

\bibitem[{Neubig and Hu(2018)}]{neubig2018rapid}
Graham Neubig and Junjie Hu. 2018.
\newblock Rapid adaptation of neural machine translation to new languages.
\newblock In \emph{Proceedings of the 2018 Conference on Empirical Methods in
  Natural Language Processing}, pages 875--880.

\bibitem[{Oren et~al.(2019)Oren, Sagawa, Hashimoto, and
  Liang}]{oren2019distributionally}
Yonatan Oren, Shiori Sagawa, Tatsunori Hashimoto, and Percy Liang. 2019.
\newblock Distributionally robust language modeling.
\newblock In \emph{Proceedings of the 2019 Conference on Empirical Methods in
  Natural Language Processing and the 9th International Joint Conference on
  Natural Language Processing (EMNLP-IJCNLP)}, pages 4218--4228.

\bibitem[{Ott et~al.(2019)Ott, Edunov, Baevski, Fan, Gross, Ng, Grangier, and
  Auli}]{ott2019fairseq}
Myle Ott, Sergey Edunov, Alexei Baevski, Angela Fan, Sam Gross, Nathan Ng,
  David Grangier, and Michael Auli. 2019.
\newblock fairseq: A fast, extensible toolkit for sequence modeling.
\newblock In \emph{Proceedings of NAACL-HLT 2019: Demonstrations}.

\bibitem[{Papineni et~al.(2002)Papineni, Roukos, Ward, and Zhu}]{Papineni:2002}
Kishore Papineni, Salim Roukos, Todd Ward, and Wei-Jing Zhu. 2002.
\newblock Bleu: a method for automatic evaluation of machine translation.
\newblock In \emph{ACL 2002}.

\bibitem[{Pires et~al.(2019)Pires, Schlinger, and
  Garrette}]{pires2019multilingual}
Telmo Pires, Eva Schlinger, and Dan Garrette. 2019.
\newblock How multilingual is multilingual bert?
\newblock In \emph{Proceedings of the 57th Annual Meeting of the Association
  for Computational Linguistics}, pages 4996--5001.

\bibitem[{Post(2018)}]{post-2018-call}
Matt Post. 2018.
\newblock \href {https://www.aclweb.org/anthology/W18-6319} {A call for clarity
  in reporting {BLEU} scores}.
\newblock In \emph{Proceedings of the Third Conference on Machine Translation:
  Research Papers}, pages 186--191, Belgium, Brussels. Association for
  Computational Linguistics.

\bibitem[{Qi et~al.(2018)Qi, Sachan, Felix, Padmanabhan, and
  Neubig}]{qi18naacl}
Ye~Qi, Devendra Sachan, Matthieu Felix, Sarguna Padmanabhan, and Graham Neubig.
  2018.
\newblock \href {https://arxiv.org/pdf/1804.06323.pdf} {When and why are
  pre-trained word embeddings useful for neural machine translation?}
\newblock In \emph{Meeting of the North American Chapter of the Association for
  Computational Linguistics (NAACL)}, New Orleans, USA.

\bibitem[{Roughgarden(2016)}]{Roughgarden16}
Tim Roughgarden. 2016.
\newblock \emph{Twenty Lectures on Algorithmic Game Theory}.
\newblock Cambridge University Press.

\bibitem[{Sagawa et~al.(2020)Sagawa, Koh, Hashimoto, and
  Liang}]{sagawa2019distributionally}
Shiori Sagawa, Pang~Wei Koh, Tatsunori~B Hashimoto, and Percy Liang. 2020.
\newblock Distributionally robust neural networks for group shifts: On the
  importance of regularization for worst-case generalization.
\newblock In \emph{International Conference on Learning Representations
  (ICLR)}, Addis Ababa, Ethiopia.

\bibitem[{Vaswani et~al.(2017)Vaswani, Shazeer, Parmar, Uszkoreit, Jones,
  Gomez, Kaiser, and Polosukhin}]{vaswani2017attention}
Ashish Vaswani, Noam Shazeer, Niki Parmar, Jakob Uszkoreit, Llion Jones,
  Aidan~N Gomez, {\L}ukasz Kaiser, and Illia Polosukhin. 2017.
\newblock Attention is all you need.
\newblock In \emph{Advances in neural information processing systems}, pages
  5998--6008.

\bibitem[{Wang et~al.(2020{\natexlab{a}})Wang, Tsvetkov, and
  Neubig}]{wang2020balancing}
Xinyi Wang, Yulia Tsvetkov, and Graham Neubig. 2020{\natexlab{a}}.
\newblock Balancing training for multilingual neural machine translation.
\newblock In \emph{Proceedings of the 58th Annual Meeting of the Association
  for Computational Linguistics}, pages 8526--8537.

\bibitem[{Wang et~al.(2020{\natexlab{b}})Wang, Lipton, and
  Tsvetkov}]{wang2020negative}
Zirui Wang, Zachary~C Lipton, and Yulia Tsvetkov. 2020{\natexlab{b}}.
\newblock On negative interference in multilingual language models.
\newblock In \emph{Proceedings of the 2020 Conference on Empirical Methods in
  Natural Language Processing (EMNLP)}, pages 4438--4450.

\bibitem[{Wang et~al.(2021)Wang, Tsvetkov, Firat, and Cao}]{wang2021gradient}
Zirui Wang, Yulia Tsvetkov, Orhan Firat, and Yuan Cao. 2021.
\newblock \href {https://openreview.net/forum?id=F1vEjWK-lH_} {Gradient
  vaccine: Investigating and improving multi-task optimization in massively
  multilingual models}.
\newblock In \emph{International Conference on Learning Representations}.

\bibitem[{Zoph et~al.(2016)Zoph, Yuret, May, and Knight}]{zoph2016transfer}
Barret Zoph, Deniz Yuret, Jonathan May, and Kevin Knight. 2016.
\newblock Transfer learning for low-resource neural machine translation.
\newblock In \emph{Proceedings of the 2016 Conference on Empirical Methods in
  Natural Language Processing}, pages 1568--1575.

\end{thebibliography}
